\documentclass{article}

\pdfoutput=1

    \usepackage[preprint]{neurips_2025}

\usepackage{longtable} 
\usepackage{graphicx} 
\usepackage{array}     
\usepackage[utf8]{inputenc}
\usepackage[T1]{fontenc}
\usepackage[dvipsnames]{xcolor}
\usepackage[utf8]{inputenc} 
\usepackage[T1]{fontenc}   
\usepackage{hyperref}      
\usepackage{booktabs}       
\usepackage{amsfonts}       
\usepackage{nicefrac}      
\usepackage{microtype}      
\usepackage{xcolor}         
\usepackage{xpatch}
\usepackage{tabularx} 
\usepackage{array} 
\usepackage{graphicx}
\usepackage{colortbl}
\usepackage{tikz}
\usepackage{multirow}
\usepackage{graphicx}
\usepackage{pgffor}
\usepackage{subcaption}
\usepackage{pifont} 
\usepackage{float}
\usepackage{algorithm} 
\usepackage{algorithmic} 
\usepackage{hyperref}
\usepackage{xpatch}
\usepackage{makecell}
\usepackage[most]{tcolorbox}
\usepackage{tcolorbox}
\usepackage{listings}
\usepackage{xcolor}
\usepackage{enumitem}
\title{CITE: A Comprehensive Benchmark for Heterogeneous Text-Attributed Graphs on Catalytic Materials}

\author{
Chenghao Zhang, Qingqing Long, Ludi Wang, Wenjuan Cui, Jianjun Yu\thanks{Corresponding Authors}, Yi Du$^*$ \\
  Computer Network Information Center, China Academy of Sciences\\
  \{chzhang,qqlong,wld,wenjuancui,yujj,duyi\}@cnic.cn \\
}
\newcolumntype{M}[1]{>{\centering\arraybackslash}m{#1}} 
\newcolumntype{L}[1]{>{\raggedright\arraybackslash}m{#1}} 

\begin{document}

\maketitle

\begin{abstract}
Text-attributed graphs(TAGs) are pervasive in real-world systems,where each node carries its own textual features. In many cases these graphs are inherently heterogeneous, containing multiple node types and diverse edge types. Despite the ubiquity of such heterogeneous TAGs, there remains a lack of large-scale benchmark datasets. This shortage has become a critical bottleneck, hindering the development and fair comparison of representation learning methods on heterogeneous text-attributed graphs. In this paper, we introduce CITE - \textbf{C}atalytic \textbf{I}nformation \textbf{T}extual \textbf{E}ntities Graph, the first and largest heterogeneous text-attributed citation graph benchmark for catalytic materials. CITE comprises over 438K nodes and 1.2M edges, spanning four relation types. In addition, we establish standardized evaluation procedures and conduct extensive benchmarking on the node classification task, as well as ablation experiments on the heterogeneous and textual properties of CITE. We compare four classes of learning paradigms, including homogeneous graph models, heterogeneous graph models, LLM(Large Language Model)-centric models, and LLM+Graph models. 
In a nutshell, we provide (i) an overview of the CITE dataset, (ii) standardized evaluation protocols, and (iii) baseline and ablation experiments across diverse modeling paradigms. 
\end{abstract}

\section{Introduction}
Graphs are frequently employed in the modeling of relationships and structures of  objects in the real world, covering a wide range of scenarios, such as citation networks~\cite{liu2013full,ju2024comprehensive}, social networks~\cite{myers2014information,long2020graph} and  recommendation systems~\cite{zhang2024influential}.
Moreover, many nodes of real world graphs are often associated with text attributes, leading to text-attributed graphs~\cite{yang2021graphformers,fang2022polarized}. In the context of TAGs, nodes are conventionally employed to represent text entities, such as documents or sentences. In addition, there are multiple types of nodes and links in the real world~\cite{wang2022molecular,long2021hgk}. For instance, in citation network, papers are connected together via papers, authors, venues, terms and keywords. Compared to homogeneous graphs, the heterogeneous graphs is capable of processing and containing more information in nodes and links and consequently establishes a new development of data mining~\cite{shi2016survey,yan2024inductive}.\newline
The crux of the learning process on TAGs is the effective integration of two key elements: the node attributes, which encompass the textual semantics, and the graph topology, which encompasses the structural connections. This integration is pivotal in facilitating the learning of node representations. In addressing graphs that encompass both node attributes and graph structural information, conventional pipelines are commonly classified into three distinct categories.

\begin{figure*}[htbp]
    \centering
    \includegraphics[width=\linewidth]{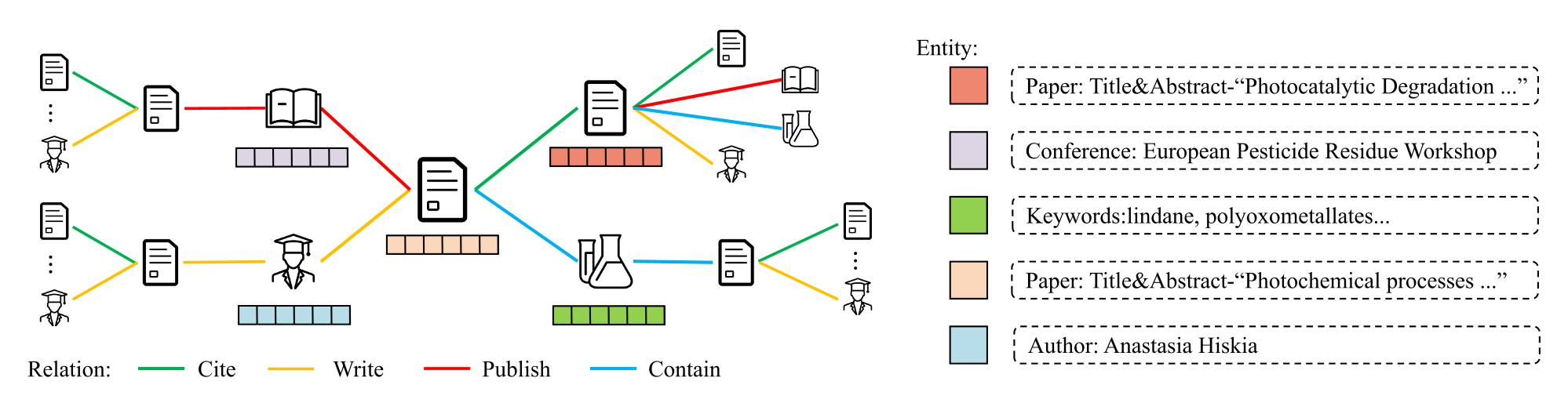} 
    \caption{Overview of CITE  Structure and Content}  
    \label{overview}  
    \vspace{-5mm}
\end{figure*}

Firstly, methods based on pretrained language models(PLMs) are effective at capturing the semantic content of textual information associated with nodes, enabling the characterization of individual node properties~\cite{yan2023comprehensive,luo2025large}. In PLM-based methods, the textual content from the target node is typically input into the model, transforming the task into a text classification or link prediction problem~\cite{chen2024exploring}. However, a significant drawback of these methods is that they neglect the topological knowledge embedded in the graph structure, which is critical due to the high non-linearity of graph topologies~\cite{zhao2022learning}.
While recent efforts have attempted to enhance LLM reasoning by explicitly injecting GNN-generated structural embeddings into LLM models' processing pipeline through techniques like graph-aware tokenization and cross-modality alignment, such approaches frequently demonstrate performance instability across different scenarios~\cite{zhang2024graphtranslator,huang2024can,tang2024graphgpt}.\newline
Secondly, methods based on Graph Neural Networks (GNNs), including homogeneous and heterogeneous graph models, can effectively capture structural relationships within a graph by leveraging the message-passing mechanism, which encodes the inherent proximity between nodes~\cite{velickovic2017graph}. However, GNNs exhibit limited capacity when handling voluminous textual information: as the amount of text associated with nodes grows, their predictive performance may actually deteriorate. To mitigate this, some approaches leverage LLMs to preprocess extensive node text and feed the resulting embeddings into GNNs, but most GNNs treat node attributes as static, unlearnable representations during message passing, hindering the backpropagation of gradients into attribute modeling. This disconnect between node attributes and graph topology impedes end-to-end training, limiting the overall effectiveness of the approach~\cite{yan2023comprehensive}. Moreover, current efforts to empower GNNs with LLM-derived representations predominantly target homogeneous graphs, while exploration in the heterogeneous setting remains comparatively scarce.\newline
Thirdly, in order to reap the benefits of both methods, recent research has illuminated the co-training paradigm, which can be called as aligner methods~\cite{zhao2022learning,jin2023patton}. The fundamental premise of the aligner methods is to enable GNNs and PLMs to facilitate the learning of topology and textual information, respectively. However, these methods can result in cumbersome models and severe scalability issues.Moreover, beyond these architectural and scalability drawbacks, the discrete nature of graph structural embeddings often conflicts with the continuous semantic space of LLMs, leading to unpredictable interactions that may result in inconsistent model behaviors depending on specific dataset characteristics or architectural implementations.\newline
Current TAG representation learning frameworks are predominantly designed under the homogeneous graph assumption, where all nodes share identical type definitions and interaction patterns. This simplification starkly contrasts with real-world graph systems that inherently manifest structural heterogeneity through diverse node/link types.  A canonical example emerges in academic citation networks, which naturally encompass multiple interacting entity types including papers, authors, journals, and research keywords, etc.
In heterogeneous text-attributed graphs (TAGs), nodes and edges exhibit type-specific semantic properties, where differentiated entity features (e.g., author affiliations, paper abstracts) and relational interaction patterns (e.g., citation-based knowledge flow, co-authorship networks) collectively mirror real-world structural complexity and domain-specific knowledge hierarchies.
Conventional homogeneous TAG methods fundamentally fail to model these type-specific attribute distributions and cross-type interaction dynamics, as they assume uniform attribute spaces and single relation propagation mechanisms.
The scarcity of dedicated heterogeneous TAG datasets has significantly hindered the advancement of representation learning in this domain, as existing methods predominantly focus on homogeneous structures that fail to reflect real-world complexity. 
Despite the existence of current databases, such as OGB~\cite{hu2020open}, Cora~\cite{sen2008collective}, PubMed~\cite{white2020pubmed}, Citeseer~\cite{giles1998citeseer}, MAG~\cite{wang2020microsoft}, they have limitations in terms of supporting heterogeneity, or leveraging text attributes. Therefore, there is now an urgent need for a heterogeneous text-attribute graph to enhance TAG representation learning, which are essential for developing and evaluating next-generation graph learning architectures capable of handling real-world heterogeneous information networks.

In order to address these discrepancies, our research proposes CITE - \textbf{C}hemical \textbf{I}nformation \textbf{T}extual \textbf{E}ntities Graph as shown in Figure\ref{overview}. CITE is a pioneering benchmark graph dataset offering a heterogeneous text-attributed citation network. Besides, we conduct extensive experiments on CITE to provide comprehensive and reliable benchmarks. In summary, our contributions are as follows:
\begin{itemize}[left=1.7em]
\item To the best of our knowledge, CITE is the first open dataset and benchmark of citation graph with a combination of heterogeneity and text-attributes.
\item We conducted a series of experiments on CITE, involving multiple classes of models, and carried out an in-depth analysis of the interaction mechanism between the model structure and the experimental results.
\item We drawn 4 key observations:(1) Heterogeneous graph models demonstrate superior robustness in comparison to homogeneous graph models. (2) The performance of LLM is impacted by misaligned predictions. (3) CITE presents new challenges with structural complexity, textual richness, and label imbalance. (4) High-degree heterogeneous nodes and rich textual semantics play a crucial role in advancing graph representation learning.
\end{itemize}

\section{Related Work}
\subsection{Academic Citation Graph}
Mccallum et.al established Cora~\cite{mccallum2000automating}, which is a scientific publication dataset of computer science. Cora contains over 50,000 papers which is categorized into seven categories: Case Based, Genetic Algorithms, Neural Networks, Probabilistic Methods, Reinforcement Learning, Rule Learning, Theory. The most used version dataset is designed for node classification, which contains 2708 scientific publications, the citation network consists of 5429 links. The nodes are implemented with one-hop encoding, where each node is a 0/1 vector of dimension 1433.
The PubMed dataset~\cite{white2020pubmed} is based on the PubMed database, designed for node classification. It contains 19,717 scientific publications related to diabetes, which are grouped into three categories:Experimental Diabetes Mellitus,  Diabetes Mellitus Type 1 and Diabetes Mellitus Type 2. The cross-reference network of these publications contained 44,338 edges. 
CiteSeer dataset~\cite{giles1998citeseer} is designed for node classification, contains 3312 scientific publications categorized into one of six categories: Agents, AI, DB, IR, ML and HCI. The citation network consists of 4732 links. Each publication in the dataset is described by a 3703-dimensional, 01-valued word vector.
OGB dataset~\cite{hu2020open} is designed for node classification, link prediction and graph property prediction. The dataset assembles graphs of various sizes and topics, e.g., proteins, product purchases, arXiv papers, etc., ranging in size from 100k to 100M.Each node has 128 dimension word2vec embedding.
Aminer citation data~\cite{tang2008arnetminer} is extracted from DBLP, ACM, MAG, and other sources. The first version contains 629,814 papers and 632,752 citations. Each paper is stored as a json file and is associated with an abstract, author, year, location, title, etc. All of the above information is in the form of strings and numbers, and there is no information about the adjacency matrix of the paper's citation graphs. 
ACM dataset~\cite{wang2019heterogeneous} is a heterogeneous graph that comprises 3025 papers, 5835 authors and 56 subjects. Paper features correspond to elements of a bag-of-words represented of keywords. The dataset is designed for link prediction task.
The Microsoft Academic Graph (MAG)~\cite{sinha2015overview} is a heterogeneous graph containing scientific publication records, citation relationships between those publications, as well as authors, institutions, journals, conferences, and fields of study. The dataset has over 150 million nodes and over 2 billion edges.

\subsection{Chemistry Graph}
In the domain of chemistry, several datasets are already available that provide a valuable resource for research and drive progress in chemistry-related tasks, especially in the realm of research related to molecular structure and properties.
Optical Chemical Structure Recognition (OCSR) is a significant direction in the field of chemical image analysis. With the aim of improving the performance of neural architectures for chemical image interpretation, Hormazabal et.al introduced a dataset which contains one million molecular images and over 700,000 curated bounding boxes for chemical entities~\cite{hormazabal2022cede}.
To improve molecular property prediction, Hamiltonian prediction, and conformational optimization tasks $\nabla^2$DFT dataset is introduced, which contains 15,716,667 conformations, including energies, forces, 17 molecular properties, Hamiltonian and overlap matrices, and a wavefunction object~\cite{khrabrov20242}.
In the field of molecular structure prediction, to addresses the limitations of single-modality approaches, Alberts et.al established a multimodal dataset of spectroscopic data involving 790k molecules. The dataset supports the development of foundation models for structure elucidation, spectra prediction, and functional group identification~\cite{alberts2024unraveling}.
Large language models (LLMs) have recently demonstrated impressive reasoning abilities across a wide array of tasks. In chemistry, Guo et.al identify three key chemistry-related capabilities of LLM, including understanding, reasoning and explaining. They also established a benchmark containing eight chemistry tasks~\cite{guo2023can}.
The Knowledge Graph (KG) is a valuable tool for integrating various information sources and providing a unified view of domain knowledge. In the domain of chemical materials, the MKG dataset has been developed, which organizes chemical entities such as "formulas," "names," and "acronyms" into a structured triad containing 162,605 nodes and 731,772 edges~\cite{ye2024construction}.
Graphs can also model compounds interaction. MOTI$V\varepsilon$  integrates morphological data and graph-based modeling by compiling Cell Painting features for 11K genes and 3.6K compounds, enhancing the drug-target interaction performance~\cite{arevalo2024motive}.

\section{CITE Dataset Overview and Construction}
\subsection{Overview of CITE}
CITE constructs a heterogeneous information network spanning the interdisciplinary domain of photocatalysis and electrocatalysis, encompassing peer-reviewed publications from 1922 to 2022, sourced from sourced from CrossRef. The network integrates four node types: Papers (research articles), Authors, Journals, and Keywords, with edges capturing semantic and relational interactions: Paper-Paper (citation relationships), Paper-Author (authorship), Paper-Journal (publication venues), Paper-Keywords (topic annotations). With 438,304 nodes and 1,220,373 edges, CITE provides a comprehensive representation of the field’s evolution, linking structural metadata (e.g., citation graphs) with rich textual attributes (e.g., abstracts, section content).
Compared to existing citation graph datasets, CITE excels in heterogeneity and semantic integration. By unifying diverse node types (papers, authors, journals, keywords) as first-class entities, it supports nuanced analyses such as institutional collaboration dynamics and keyword-driven topic evolution, which are capabilities absent in homogeneous citation networks. Unlike prior works limited to metadata, CITE embeds abstract content  and keywords directly into graph nodes, enabling cross-modal queries that link methodological details to citation patterns, providing a robust foundation for longitudinal studies. As quantified in Table\ref{comparsion graphs},
CITE surpasses benchmarks in node diversity and attribute richness.
CITE demonstrates balanced distributions across nodes, edges, and semantic labels, addressing common skewness issues in domain-specific graphs, avoiding dominance by a single entity and ensuring robust representation of heterogeneous interactions. 
For more detailed statistical information on the composition of CITE and its boarder impacts, please refer to the Appendix\ref{appendixA}.\newline
It is worth noting that journals and keywords collectively represent merely 12.4\% of nodes but account for 31.5\% of total edges. This structural characteristic suggests two critical insights: 
(1) Research topics exhibit natural concentration around photocatalysis and electrocatalysis domains as manifested through keyword co-occurrence patterns, and (2) Journal entities, despite constituting merely 0.35\% of nodes, account for 10.5\% of edges (paper-journal affiliations), indicating concentrated paper dissemination where a limited subset of journals structurally anchor majority of publications. \newline
CITE generates structured CSV files by processing metadata to support efficient graph construction and downstream tasks. The resulting file includes core fields such as id, DOI, title, year, and author, with detailed information about these fields and their interpretations provided in Appendix\ref{appendixA}.
CITE is publicly hosted on huggingface
, including raw JSON files metadata and processed CSV files. This structured approach ensures CITE’s utility in diverse tasks, from citation prediction to interdisciplinary trend analysis, while maintaining rigorous reproducibility standards.

\begin{table}[]

\centering
\setlength{\tabcolsep}{4pt}
\caption{Statistics of graphs, where NC presents Node classification, LP presents Link prediction}
 \resizebox{\linewidth}{!}{
\begin{tabular}{ccccccccc}
\bottomrule
\textbf{Dataset}    & \textbf{Nodes}    & \textbf{Edges}     & \textbf{Class} & \textbf{Modeling} & \textbf{Node types} & \textbf{Hetero.} & \textbf{Task} & \textbf{Raw-text} \\ \midrule
Cora~\cite{mccallum2000automating}       & 2708      & 5,429     & 7     & One-hop  & 1       & \textcolor{red}{\ding{55}}      & NC & \textcolor{red}{\ding{55}} \\ 
Citeseer~\cite{giles1998citeseer}   & 3,312     & 4,732     & 6     & One-hop  & 1       & \textcolor{red}{\ding{55}}      & NC  & \textcolor{red}{\ding{55}} \\  
Pubmed~\cite{white2020pubmed}     & 19,717    & 44,338    & 3     & Tf-IDF   & 1       & \textcolor{red}{\ding{55}}      & NC  & \textcolor{red}{\ding{55}} \\  
Ogbn-arxiv~\cite{hu2020open} & 169,343   & 1,166,243 & 40    & Word2vec & 1       & \textcolor{red}{\ding{55}}      & NC  & \textcolor{green}{\ding{51}} \\  
ACM~\cite{wang2019heterogeneous}        & 8860 & -         & 56    & BoW & 2         & \textcolor{green}{\ding{51}}  & LP    & \textcolor{red}{\ding{55}} \\  
CITE       & 438,304     & 1,220,373         & 85   & Bert     & 4   & \textcolor{green}{\ding{51}}  & NC  & \textcolor{green}{\ding{51}} \\ \bottomrule
\end{tabular}
}
\label{comparsion graphs}
\end{table}
\subsection{Collection and Construction of CITE}
CITE is derived from peer-reviewed publications in the interdisciplinary domains of photocatalysis and electrocatalysis, sourced from crossref, which encompasses critical journals and conferences in catalysis, such as Applied Catalysis B: Environmental, ACS Catalysis, and Electrochimica Acta. The temporal scope spans 1922 to 2022, a period capturing the evolution of the field from foundational studies on TiO2-based photocatalysts to modern advances in atomic-scale electrocatalyst design. The raw data comprises complementary components essential for constructing the heterogeneous graph:\newline Structured Metadata is extracted from Crossref database, which includes fundamental bibliographic attributes such as paper titles, author lists, and journal/conference names, etc.
In order to obtain relevant literature in the field of electrocatalytic, we firstly found the dois of scientific literature through crossref. Specifically, we exported metadata from over 150k articles from crossref using the keywords "photocataly*" and "electrocataly*" as subject indexes. The curated metadata is structured into JSON files organized by paper DOI, where each entry integrates DOI number, title of the paper, year of publication, first author, all authors' information, etc., facilitating downstream graph construction. Please refer to Appendix\ref{appendixA} for the metadata crawling, disambiguation and organization process and the example of the curated metadata.\newline
CITE consists of four core entity types: Paper, Author, Journal, and Keywords, which collectively correspond to four types of relationships: Paper-Paper, Paper-Author, Paper-Journal, and Paper-Keywords.
Paper nodes represent article content, with attributes like title and abstract. Labels are assigned based on the SCI classification of the journal, or 'uncategorized' if not indexed. Author nodes represent article authors, identified by ORCID or, in the absence of ORCID, by institutional affiliation if names match. Journal nodes represent publishing journals or conferences, with their ISSNs extracted from metadata for official title conversion. Keyword nodes reflect the core research topics and are extracted from paper metadata.To ensure data uniqueness, each node is assigned a unique identifier. Edges capture explicit relationships (citations, authorship) and implicit semantic connections (topical similarity).  Citation edges (Paper-Paper) are based on Crossref metadata references. Authorship edges (Paper-Author) are derived from author lists, with each author uniquely identified. Publication venue edges (Paper-Journal) link papers to journals via ISSN. Topical relevance edges (Paper-Keywords) are created from paper keywords, with data cleaning applied.
\section{Experiment}
CITE is a newly-proposed heterogeneous textual attribute graph for scientific literature analysis in catalytic materials, characterized by its integration of multi-type nodes and rich text attributes. It possesses unique domain-specific characteristics and compound heterogeneity-textuality properties. According to our observation, modeling compound heterogeneity-textuality structures introduces unique challenges in semantic interaction patterns, multi-granularity integration, and architectural adaptability across learning paradigms.
To systematically validate the value of this dataset, we design experiments around three core questions:\newline
\textbf{Q1: What is the impact of heterogeneous text-attributed graph architectures on model generalization across diverse learning paradigms?}  \newline 
\textbf{Q2: Does CITE present unique challenges compared to existing citation graph benchmarks that cannot be addressed by current methods?} \newline
\textbf{Q3: Are multi-relational structures and textual semantic richness necessary conditions for robust graph representation learning?} \newline
To systematically address these research questions, we employ four learning paradigms: (1) 
Homogeneous graph models, (2) Heterogeneous graph models, (3) LLM-centric models (4) LLM+Graph models.
All experiments follow a 60\%/20\%/20\% split for training/validation/test, and all graph models share BERT-base tokenization with 128-token truncation.
Our study centers on graph-based node classification for scientific paper categorization, specifically addressing the prediction of discipline-specific labels from the SCI taxonomy (such as "Electrochemical Engineering" and "Matrials Chemistry") through heterogeneous academic network analysis. We evaluate performance using Macro-F1 and Micro-F1, with model-specific hyperparameters (e.g., learning rates),error bars and execution information detailed in Appendix\ref{appendixB}.

\subsection{Baseline Model Comparison \textit{(Q1)}}
\textbf{Experiment Setup} \newline
To systematically assess how heterogeneous text-attributed graph architectures impact model generalization across learning paradigms, we evaluate the learning paradigms above. 
For traditional graph models, we test homogeneous architectures (GCN~\cite{kipf2016semi}, GAT~\cite{velickovic2017graph} , GraphSAGE~\cite{hamilton2017inductive}) and heterogeneous counterparts (RGCN~\cite{schlichtkrull2018modeling}, CompGCN~\cite{vashishth2019composition}, SimpleGCN~\cite{lv2021we}, HPN~\cite{ji2021heterogeneous}, HGT~\cite{hu2020heterogeneous}, NARS~\cite{yu2020scalable}) to isolate the impact of modeling multi-type nodes and edges. As LLM-centric models, we employ LLaMA-7B model.
To explore the synergy between textual attributes and graph structures, we evaluate LLM+Graph models, including TAPE~\cite{he2023harnessing} (LLM enhance GNN in prediction) and GraphGPT~\cite{tang2024graphgpt} (graph construction enhance LLM in prediction). 
For architectures insensitive to structural heterogeneity (homogeneous graph models and models involving LLMs), we implement targeted adaptations: For homogeneous graph models we discard explicit heterogeneous nodes (authors/keywords/journals) while preserving their semantic influence by fusing multi-typed information into paper node features, then encoded via BERT. This creates a homogeneous graph containing only paper nodes but with enriched textual features.
For models involving LLMs, we implement structured prompt engineering that injects heterogeneous metadata: paper nodes are contextualized with connected authors, journals, and keywords through templated descriptors, with full templates detailed in Appendix\ref{appendixC}. 
All models share identical BERT text encoders, training splits and evaluation metrics, ensuring fair comparison of paradigm-specific generalization capabilities.
Experiment results are shown in Table\ref{Q1} and the top three results for each metric are highlighted in \colorbox{red!60}{red}, \colorbox{yellow!60}{yellow}, and \colorbox{cyan!60}{blue}. 

\textbf{Discussion}\newline
\textbf{(1) Weaknesses in the perception of structure hinder the performance of homogeneous graph models}: 
Results reveal critical insights into architectural efficacy on heterogeneous text-attributed graphs. Structural heterogeneity emerges as a decisive performance driver: heterogeneous graph models vastly outperform homogeneous counterparts and LLM-centric model, underscoring the necessity of explicit type-aware modeling for scientific categorization. The significant decline in Macro-F1 performance in homogeneous adaptations stems from semantic dilution during node-type compression: flattening authors, journals, and keywords into paper features erases disciplinary signals critical for rare classes (e.g., only 0.6\% "Crystallography" labels). This degradation intensifies under CITE’s long-tailed label distribution, where homogeneous graph models’ isotropic aggregation fails to prioritize journal nodes which are strong discipline indicators.\newline
\textbf{(2) GNN-based LLM+Graph models architectures exhibit enhanced robustness}: 
LLM+Graph models exhibit duality: 
TAPE employs a two-stage cascaded framework: it first uses LLM to predict pseudo-labels from paper text prompts, then injects these probabilistic pseudo-labels as node features into a graph model for structural refinement. This decoupling mitigates LLMs' answer stochasticity—by treating LLM outputs as noisy-but-informative priors rather than final predictions, TAPE achieves robust Micro-F1 despite CITE's extensive taxonomy.
On the other hand, GraphGPT explores an end-to-end paradigm where graph embeddings encoded by graph models serve as input to LLM for text-based classification. In our experimental setting with extensive discipline categories, this approach faces challenges in maintaining output consistency, generating class name variants (e.g., expanding "Engineering" to "Engineering Studies") or producing null responses under complex taxonomic constraints, so the performance of this approach was found to be unsatisfactory. This indicates challenges in precise label alignment when mapping graph-informed LLM outputs to predefined taxonomic categories.\newline
\textbf{(3) Substandard consistency of LLM models' output}: 
Both LLM-centric and LLM+Graph models exhibit inherent limitations in extreme multi-class scenarios, with LLM-driven outputs showing non-deterministic tendencies across paradigms as detailed in Appendix\ref{appendixC}. However, TAPE’s decoupled design introduces error-correction mechanisms: the GNN layer refines initial LLM predictions through structural consensus, such as resolving ambiguous "MATERIALS\_SCI" classifications by reinforcing journal-specific patterns. In contrast, GraphGPT’s tightly-coupled pipeline, while innovative in bridging graph-text spaces, demonstrates compounding sensitivity to semantic ambiguities when handling fine-grained labels.
Notably, even the best heterogeneous graph models and LLM+Graph models show room for improvement, 0.5 Micro-F1 in average indicates persistent minority-class struggles, likely due to edge-type imbalance (e.g., sparse cross-disciplinary citations). 
\begin{tcolorbox}[colback=gray!5!white, colframe=gray!80!black, title=Summary.,height=1.8cm]
Compared to homogeneous graph models, heterogeneous graph models exhibit greater robustness.
Models related to LLM struggle with accurate label alignment.
\end{tcolorbox}

\begin{table}[htbp]
\caption{Performance Across Learning Paradigms on CITE}
\centering
\resizebox{0.8\linewidth}{!}{
\small
\begin{tabular}{ccccc}
\toprule
\textbf{Category}  & \textbf{Model Name} & \textbf{Micro-f1} & \textbf{Macro-f1}  \\ \midrule
\multirow{3}{*}{\begin{tabular}[c]{@{}c@{}} Homogeneous Graph Model\end{tabular}}   & GCN & 0.5057& 0.0239    \\
 & GAT   & 0.5115   & 0.0457   \\
 & GraphSAGE  & 0.5117   & 0.0474   \\ \midrule
\multirow{8}{*}{\begin{tabular}[c]{@{}c@{}}Heterogeneous Graph Model\end{tabular}} & RGCN  & 0.8626   & 0.1442  \\
 & CompGCN    & 0.8173   & 0.2184   \\
 & SimpleHGN  & \colorbox{yellow!60}{0.9907}  & \colorbox{yellow!60}{0.5052}   \\
   & HPN   & 0.7863   & 0.1387  \\
  & HGT   & \colorbox{cyan!60}{0.9791}   & \colorbox{cyan!60}{0.4183}  \\
 & NARS  & 0.7273   & 0.0660  \\ 
 & HAN & 0.7515 & 0.1706\\
 & MAGNN &0.5542&0.0268\\
 \midrule
LLM-centric Model  & LLaMA & 0.0037   & 0.0614   \\ \midrule
\multirow{2}{*}{LLM+Graph Model}  & Graphgpt(Vicuna 7B) & 0.0214 & 0.0039 \\
 & TAPE(LLaMA 7B)      & \colorbox{red!60}{0.9936}   & \colorbox{red!60}{0.5655}  \\ \bottomrule
\end{tabular}
}
\label{Q1}
\end{table}

\subsection{Comparison with Existing Datasets \textit{(Q2)}}

\textbf{Experiment Setup} \newline
To address Q2, we conduct cross-domain evaluations comparing model performance on specialized benchmarks against our CITE dataset. We select four orthogonal benchmark categories: (1) homogeneous non-text-attributed graphs (arXiv, Cora, PubMed)~\cite{li2024glbench}, representing traditional structural modeling tasks without text; (2) homogeneous text-attributed graphs (Ogbn-arxiv~\cite{hu2020open}, Cora~\cite{chen2024exploring}, Pubmed~\cite{chen2024exploring}), where nodes have rich textual features but lack relational diversity; (3) heterogeneous non-text-attributed graphs (ACM, DBLP)~\cite{han2022openhgnn}, focusing on multi-relational structures without textual semantics; and (4) CITE-our heterogeneous text-attributed graph dataset, combining both challenges. See more details about the datasets in Appendix\ref{appendixC}.
For models, we selected some of the homogeneous graph models in Q1 and all the other models, they are paired with their "native" benchmarks: homogeneous graph models on arXiv/Cora/PubMed, heterogeneous graph models on ACM/DBLP, and LLM-centric/LLM+Graph models on Ogbn-arXiv/Cora/Pubmed full-text version. 
All models share identical hyperparameters, text encoders and datas split rates from Q1. Performance is measured via Micro-F1 and Macro-F1.
The results are shown in Figure\ref{different datasets} and detailed information in Appendix\ref{appendixC}.

\textbf{Discussion}\newline
\textbf{(1) Long-Tail challenges in heterogeneous-textual learning}:
The pronounced divergence between Micro-F1 and Macro-F1 scores across nearly all models, which reveals the pervasive influence of class imbalance in CITE. This phenomenon is particularly severe in models designed for homogeneous graph models or LLM-centric models, where Macro-F1 scores on CITE plummet to near-zero values, suggesting that the compound heterogeneity-textuality property exacerbates long-tail challenges. This limitation is critical in real-world applications like new material discovery and classification, where minority classes often carry high practical significance. Addressing this imbalance necessitates novel techniques, such as hybrid sampling strategies or loss reweighting.\newline
\textbf{(2) Failure in joint modeling of heterogeneous relations and textual semantics}:
Current architectures struggle to harmonize multi-relational structures with textual semantics, exposing a fundamental gap in joint modeling capabilities. Homogeneous graph models, while effective on single-type graphs, fail catastrophically on CITE due to their inability to leverage cross-type node semantics: author affiliations, journal contexts and keyword semantics are collapsed into monolithic node features. LLM-centric models perform poorly on CITE compared to other datasets. Despite its graph-aware pretraining, GraphGPT is also unsatisfactory, suggesting that even advanced LLMs still have significant potential in inferring implicit graph topology from textual prompts. Even LLM+Graph models like TAPE, though achieving high Micro-F1 on CITE through GNN-enhanced LLMs, rely on ad-hoc homogenization of heterogeneous data, which inherently distorts relational semantics. This workaround, while effective for simple heterogeneity, becomes unsustainable in complex scenarios requiring explicit multi-relational reasoning (e.g., differentiating "author-paper" collaborations from "paper-keywords" topical links). The success of heterogeneous grph models partially mitigates this issue but falters when textual richness escalates, as their rigid message-passing frameworks cannot fully exploit unstructured text. These observations highlight an urgent need for architectures that natively unify textual and structural signals without sacrificing relational granularity.\newline
\textbf{(3) Metric discrepancy and the imperative for holistic evaluation}:
The stark contrast between Micro-F1 and Macro-F1 on CITE  reflects the inadequacy of conventional metrics in capturing class imbalance. High Micro-F1 values signal overfitting to majority classes (e.g., well-represented disciplines like "Materials Chemistry"), while relatively low Macro-F1 scores expose systemic neglect of tail classes (e.g., niche fields like "Crystallography"). 
Traditional evaluation frameworks, fail to account for this complexity, risking inflated performance claims and misaligned real-world utility. To address this, future work must develop task-aware evaluation protocols that: Incorporate distribution-sensitive metrics, Weight classes by domain-specific importance.
\begin{tcolorbox}[colback=gray!5!white, colframe=gray!80!black, title=Summary.,height=1.8cm]
CITE imposes unique challenges on structural complexity, textual richness, and label imbalance.
\end{tcolorbox}

\begin{figure}[htbp]
    \centering
    \begin{minipage}{0.325\textwidth}
        \centering
        \includegraphics[width=\linewidth]{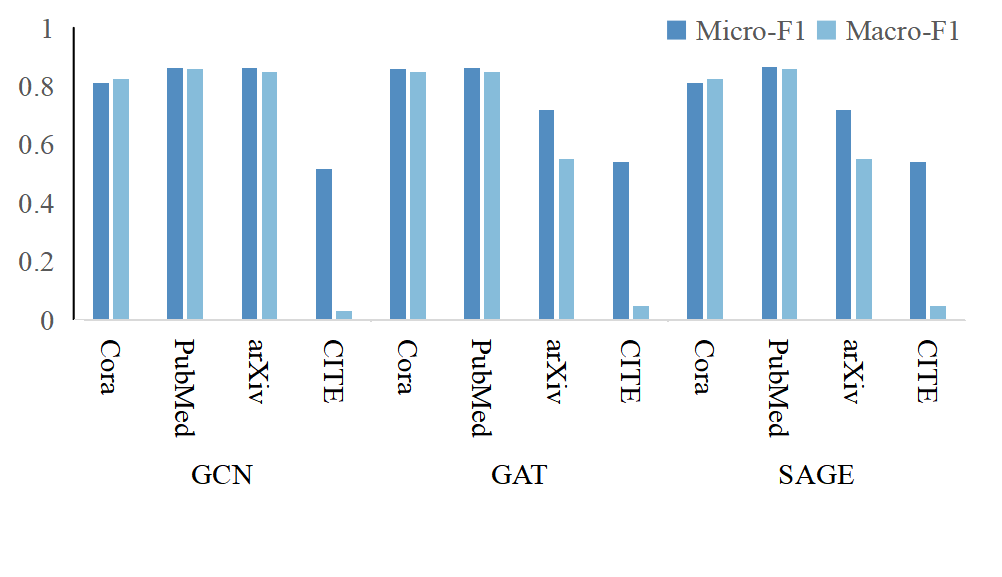}
        \subcaption{Homogeneous Graph Models}\label{fig4a}
    \end{minipage}
    \begin{minipage}{0.325\textwidth}
        \centering
        \includegraphics[width=\linewidth]{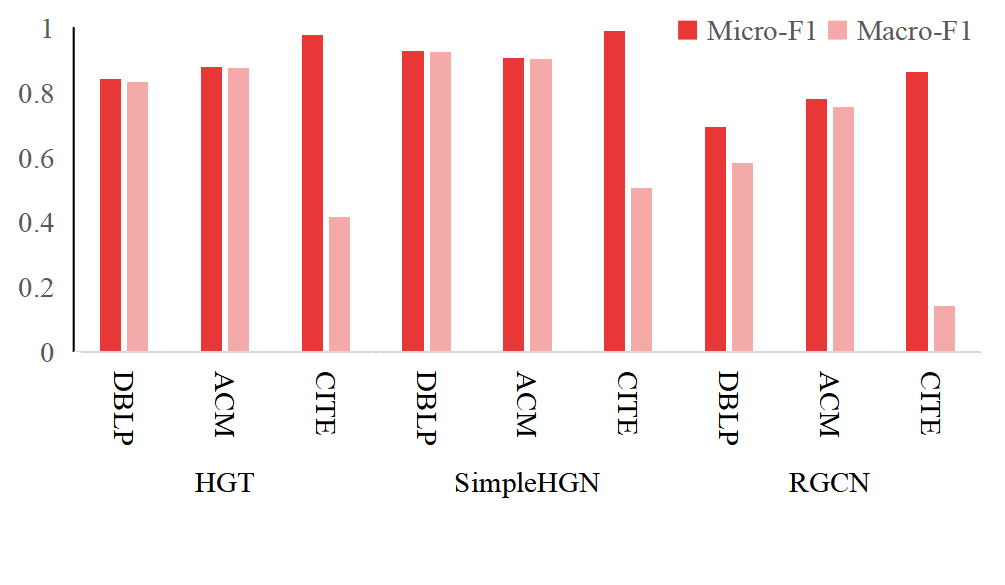}
        \subcaption{Heterogeneous Graph Models}\label{fig4b}
    \end{minipage}
    \begin{minipage}{0.325\textwidth}
        \centering
        \includegraphics[width=\linewidth]{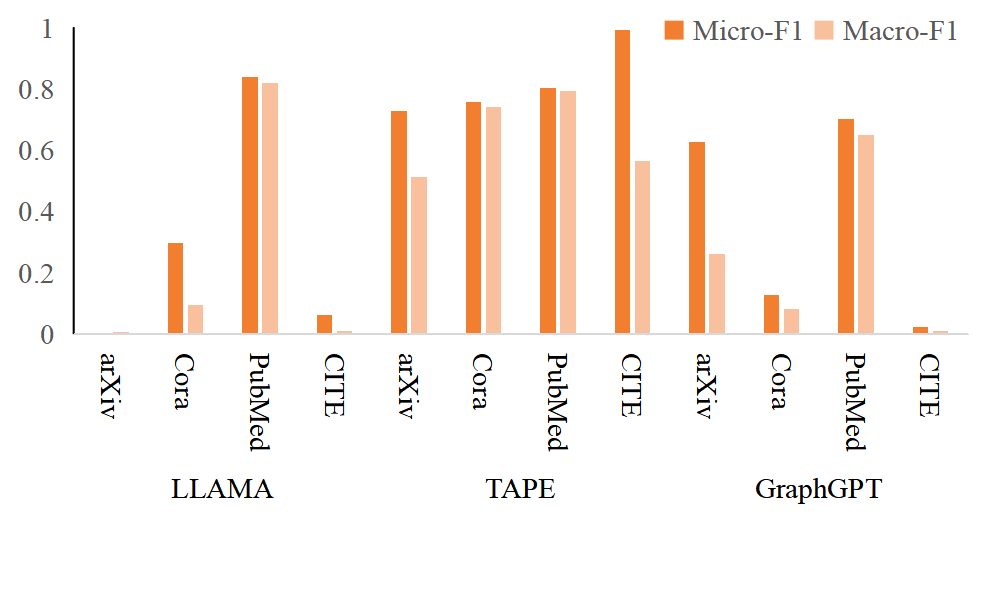}
        \subcaption{LLM and LLM+Graph models}\label{fig4c}
    \end{minipage}
    \caption{Model Performance Across Different Datasets}
    \label{different datasets}
\end{figure}

\subsection{Ablation Study on Heterogeneous Properties \textit{(Q3)}}

\textbf{Experiment Setup}\newline
To address Q3, we  designed three ablation dimensions: removal of structural heterogeneity, reduction of textual features, and elimination of both. 
In the part of heterogeneity removal, we systematically remove specific node types and their relationship with paper node (author, keywords, journal). For removing all non-paper nodes, we retain only paper nodes and paper-paper edges.
For text attribute ablation, we remove abstract content from paper nodes, leaving titles and heterogeneous structures.
In terms of full homogenization, we combine structural and textual ablations and remove all non-paper nodes and abstract text, resulting in a homogeneous graph with paper nodes (title-only text) and paper-paper edges.
For model selection, we select GCN, GAT, SAGE as homogeneous graph models, HGT, SimpleHGN, RGCN as heterogeneous graph models, LLaMA as LLM-centric model and TAPE as LLM+Graph model.
Hyperparameters and text encoders settings of models, data split ratios are the same as in Q1. Performance is measured via Micro-F1 and Macro-F1.
Experiment results are shown in Table\ref{4.3}, the best results of each model are in \textbf{bold} and the worst are \underline{underlined}.

\textbf{Discussion}\newline
\textbf{(1) Incorporation of structural information facilitates graph representation learning}: 
In homogeneous graph models, the performance of models upon removing heterogeneous information increased slightly  
and it increased more upon removing textual attributes 
reveals their inability of handling large amount of textual information. 
However the models achieved the optimal value when removing the text features and retaining the heterogeneous features, this suggests homogeneous graph models can benefit from controlled heterogeneous semantic fusion,
but fail to disentangle structural and textual signals effectively. In terms of the performance of the heterogeneous graph models, the modeling performance will degrade regardless of which heterogeneous node is eliminated, and this is especially noticeable on the journal node. When journal nodes are removed the performance drop a lot
exceeding losses from author or keywords ablation by far, which reveals low-frequency (0.58\%) but high-connectivity (average degree=50) nodes can form latent topic clusters.\newline
\textbf{(2) Rich semantics support model understanding of graphs}: 
LLaMA’s performance increase slightly 
under structural ablation and increase more 
under text attributes ablation. When making the graph full Homogenization, the results is between both of them, indicating model performance indifference to graph topology.
Also we can see the more heterogeneous nodes is integrated, the more degrade performance have, likely because discontinuous entity mentions disrupt coherent context. Unlike GNNs, LLMs treat nodes as isolated text snippets, failing to leverage relational dependencies. However, the model reached the best performance when all structural is truncated 
, which reveals that textual featrue can bring more enhancement on task performance and LLMs have potential on handling these features.\newline
\textbf{(3) LLM+Graph architecture is capable of comprehending both crucial graph structures and textual semantics}:
Despite TAPE’s overall robustness
, its dual sensitivity patterns reveal underappreciated modality interactions. First, textual sensitivity manifests in the lowest Micro-F1 of all ablation strategies when abstracts are removed, attributable to its LLM model component's reliance on discourse continuity. Paradoxically, removing author nodes improves performance slightly, which is likely because pruning low-information entities (e.g., authors with generic affiliations) elevates textual signal-to-noise ratio. Second, structural awareness emerges in the Macro-F1 decline upon  journal node ablation, contrasting with gains from removing keywords or authors. This inversion aligns with journal nodes’ high average degree, where their removal fragments latent disciplinary pathways. Collectively, these results underscore the complementary roles of heterogeneous structural cues and textual semantics in graph representation learning. Future work should focus on fully harnessing their synergistic potential.

\begin{tcolorbox}[colback=gray!5!white, colframe=gray!80!black, title=Summary., height=1.8cm]
High-degree heterogeneous nodes, rich textual semantics are important for graph representation learning.
\end{tcolorbox}

\begin{table}[]
\caption{Experiment results of ablation study (Micro-F1/Macro-F1)}
 \resizebox{\linewidth}{!}{
\begin{tabular}{ccccccccc}
\toprule
\multicolumn{1}{c}{\multirow{2}{*}{\textbf{Model Type}}}                  & \multirow{2}{*}{\textbf{Model Name}} & \multirow{2}{*}{\textbf{Original Graph}} & \multicolumn{4}{c}{\textbf{\makecell{Remove Heterogeneous Nodes}}}       & \multirow{2}{*}{\textbf{\makecell{Remove \\Text Attributes}}} & \multirow{2}{*}{\textbf{\makecell{Full \\Homogenization}}} \\ \cline{4-7}
\multicolumn{1}{c}{}  &  &   & Keywords  & Journal& Author & All &  &   \\ \midrule
\multicolumn{1}{c}{\multirow{3}{*}{\textbf{\makecell{Homogeneous\\Graph Models}}}}     & \textbf{GCN} & \underline{0.5057}/0.0239  & 0.5164/0.0294 & 0.5074/0.0243 & 0.5098/0.0261 & 0.5088/\underline{0.0186} & \textbf{0.5295/0.0407}  & 0.5084/\underline{0.0186}  \\
\multicolumn{1}{c}{}    & \textbf{GAT} & \underline{0.5115}/0.0457 & 0.5262/0.0578 & 0.5139/0.0524 & 0.5237/0.0564 & 0.5121/0.0389 & \textbf{0.5879/0.1146} & 0.5123/\underline{0.0370}  \\
\multicolumn{1}{c}{}  & \textbf{SAGE}     & 0.5117/0.0474   & 0.5254/0.0572 & 0.5157/0.0499 & \underline{0.5099}/\underline{0.0261} & 0.5113/0.0363 & \textbf{0.5873/0.1036}  & 0.5123/0.0378   \\ \midrule
\multicolumn{1}{c}{\multirow{4}{*}{\textbf{\makecell{Heterogeneous\\Graph Models}}}}   & \textbf{SimpleHGN} & 0.9907/0.5052  & 0.9904/0.4913 & 0.5471/0.0810 & 0.9881/0.4707 & 0.5027/0.0399 & \textbf{0.9913/0.5224} & \underline{0.4995}/\underline{0.0342}\\
\multicolumn{1}{c}{}  & \textbf{HGT}  & \textbf{0.9791}/0.4183 & 0.9782/\textbf{0.4253} & 0.5389/0.0613 & 0.9743/0.3915 & 0.5260/0.0646 & 0.9770/0.4094   & \underline{0.5154}/\underline{0.0554}\\
\multicolumn{1}{c}{}  & \textbf{RGCN} & 0.8626/0.1442  & 0.8398/0.936 & 0.5128/\textbf{0.1756} & 0.8317/0.0940 & 0.4978/\underline{0.0127} & \textbf{0.8637}/0.1231   & \underline{0.4971}/\underline{0.0127}     \\
\multicolumn{1}{c}{}  & \textbf{NARS} & 0.7273/0.0660   & 0.7214/0.0650 & \underline{0.4980}/\underline{0.0121} & 0.7290/0.0661 & \underline{0.4980}/\underline{0.0121} & \textbf{0.7403/0.0674}  & \underline{0.4980}/\underline{0.0121}  \\
\midrule
\multicolumn{1}{c}{\textbf{LLM-Centric Model}}& \textbf{LLaMA} & \underline{0.614}/\underline{0.0037}  & 0.0634/0.0038 & 0.0696/0.0041 & 0.0627/0.0038 & \textbf{0.1316/0.0110} & 0.0774/0.0046  & 0.0712/0.0041 \\ \midrule
\multicolumn{1}{c}{{\textbf{\makecell{LLM+Graph Model}}}} & \textbf{TAPE}   & 0.9936/0.5655  & 0.9937/0.5880 & 0.9912/\underline{0.5433} & \textbf{0.9940/0.5904} & 0.9926/0.6394 & \underline{0.9907}/0.5498& 0.9908/0.5606 \\

\bottomrule
\end{tabular}
}
\label{4.3}
\end{table}

\section{Conclusions}\label{chapter5}
In this paper, we propose CITE, a citation graph benchmark for catalytic materials. CITE successfully alleviates the challenges of limited dataset scale, including single node and relationship types and limited textual information
We have collected a hundred years of papers on photocatalysis and electrocatalysis in crossref and methodically organized them into a series of paper citation graphs, which depict four distinct types of nodes and four types of edges. We then evaluated the model on a variety of learning paradigms to explore the challenges that heterogeneous attributes and text-attributes pose to existing learning paradigms and future research directions. Finally we have organized our implemented code so that researchers can easily replicate our experiments and evaluate new models using CITE. 
\newpage
\bibliographystyle{Ref} 
\small
\bibliography{Reference}

\begin{thebibliography}{46}
\providecommand{\natexlab}[1]{#1}
\providecommand{\url}[1]{\texttt{#1}}
\expandafter\ifx\csname urlstyle\endcsname\relax
  \providecommand{\doi}[1]{doi: #1}\else
  \providecommand{\doi}{doi: \begingroup \urlstyle{rm}\Url}\fi

\bibitem[Alberts et~al.(2024)Alberts, Schilter, Zipoli, Hartrampf, and
  Laino]{alberts2024unraveling}
Alberts , M., Schilter , O., Zipoli , F., Hartrampf , N., \& Laino , T. (2024)
\newblock Unraveling molecular structure: A multimodal spectroscopic dataset
  for chemistry.
\newblock \emph{Advances in Neural Information Processing Systems} {\bfseries
  37}:\penalty0 125780--125808.

\bibitem[Arevalo et~al.(2024)Arevalo, Su, Carpenter, and
  Singh]{arevalo2024motive}
Arevalo , J., Su~, E., Carpenter , A., \& Singh , S. (2024)
\newblock Motive: A drug-target interaction graph for inductive link
  prediction.
\newblock \emph{Advances in Neural Information Processing Systems} {\bfseries
  37}:\penalty0 140320--140333.

\bibitem[Chen et~al.(2024)Chen, Mao, Li, Jin, Wen, Wei, Wang, Yin, Fan, Liu,
  et~al.]{chen2024exploring}
Chen , Z., Mao , H., Li~, H., Jin , W., Wen , H., Wei , X., Wang , S., Yin ,
  D., Fan , W., Liu , H., \& others  (2024)
\newblock Exploring the potential of large language models (llms) in learning
  on graphs.
\newblock \emph{ACM SIGKDD Explorations Newsletter} {\bfseries 25}\penalty0
  (2):\penalty0 42--61.

\bibitem[Fang et~al.(2022)Fang, Xu, Song, Long, and Zhang]{fang2022polarized}
Fang , Z., Xu~, L., Song , G., Long , Q., \& Zhang , Y. (2022)
\newblock Polarized graph neural networks. In
\newblock \emph{Proceedings of the ACM web conference 2022}
\newblock pages 1404--1413.

\bibitem[Giles et~al.(1998)Giles, Bollacker, and Lawrence]{giles1998citeseer}
Giles , C.~L., Bollacker , K.~D., \& Lawrence , S. (1998)
\newblock Citeseer: An automatic citation indexing system. In
\newblock \emph{Proceedings of the third ACM conference on Digital libraries}
\newblock pages 89--98.

\bibitem[Guo et~al.(2023)Guo, Nan, Liang, Guo, Chawla, Wiest, Zhang,
  et~al.]{guo2023can}
Guo , T., Nan , B., Liang , Z., Guo , Z., Chawla , N., Wiest , O., Zhang , X.,
  \& others  (2023)
\newblock What can large language models do in chemistry? a comprehensive
  benchmark on eight tasks.
\newblock \emph{Advances in Neural Information Processing Systems} {\bfseries
  36}:\penalty0 59662--59688.

\bibitem[Hamilton et~al.(2017)Hamilton, Ying, and
  Leskovec]{hamilton2017inductive}
Hamilton , W., Ying , Z., \& Leskovec , J. (2017)
\newblock Inductive representation learning on large graphs.
\newblock \emph{Advances in neural information processing systems} {\bfseries
  30}.

\bibitem[Han et~al.(2022)Han, Zhao, Yang, Zhang, Liu, Wang, and
  Shi]{han2022openhgnn}
Han , H., Zhao , T., Yang , C., Zhang , H., Liu , Y., Wang , X., \& Shi , C.
  (2022)
\newblock Openhgnn: An open source toolkit for heterogeneous graph neural
  network. In
\newblock \emph{Proceedings of the 31st ACM International Conference on
  Information \& Knowledge Management}
\newblock pages 3993--3997.

\bibitem[He et~al.(2023)He, Bresson, Laurent, Perold, LeCun, and
  Hooi]{he2023harnessing}
He~, X., Bresson , X., Laurent , T., Perold , A., LeCun , Y., \& Hooi , B.
  (2023)
\newblock Harnessing explanations: Llm-to-lm interpreter for enhanced
  text-attributed graph representation learning.
\newblock \emph{arXiv preprint arXiv:2305.19523}

\bibitem[Hormazabal et~al.(2022)Hormazabal, Park, Lee, Han, Jo, Lee, Jo, Kim,
  Choo, Lee, et~al.]{hormazabal2022cede}
Hormazabal , R., Park , C., Lee , S., Han , S., Jo~, Y., Lee , J., Jo~, A., Kim
  , S.~H., Choo , J., Lee , M., \& others  (2022)
\newblock Cede: A collection of expert-curated datasets with atom-level entity
  annotations for optical chemical structure recognition.
\newblock \emph{Advances in Neural Information Processing Systems} {\bfseries
  35}:\penalty0 27114--27126.

\bibitem[Hu et~al.(2020{\natexlab{a}})Hu, Fey, Zitnik, Dong, Ren, Liu, Catasta,
  and Leskovec]{hu2020open}
Hu~, W., Fey , M., Zitnik , M., Dong , Y., Ren , H., Liu , B., Catasta , M., \&
  Leskovec , J. (2020.
\newblock {\natexlab{a}})
\newblock Open graph benchmark: Datasets for machine learning on graphs.
\newblock \emph{Advances in neural information processing systems} {\bfseries
  33}:\penalty0 22118--22133.

\bibitem[Hu et~al.(2020{\natexlab{b}})Hu, Dong, Wang, and
  Sun]{hu2020heterogeneous}
Hu~, Z., Dong , Y., Wang , K., \& Sun , Y. (2020.
\newblock {\natexlab{b}}) Heterogeneous graph transformer. In
\newblock \emph{Proceedings of the web conference 2020}
\newblock pages 2704--2710.

\bibitem[Huang et~al.(2024)Huang, Han, Yang, Bao, Tao, Chai, and
  Zhu]{huang2024can}
Huang , X., Han , K., Yang , Y., Bao , D., Tao , Q., Chai , Z., \& Zhu , Q.
  (2024)
\newblock Can gnn be good adapter for llms?. In
\newblock \emph{Proceedings of the ACM Web Conference 2024}
\newblock pages 893--904.

\bibitem[Ji et~al.(2021)Ji, Wang, Shi, Wang, and Yu]{ji2021heterogeneous}
Ji~, H., Wang , X., Shi , C., Wang , B., \& Yu~, P.~S. (2021)
\newblock Heterogeneous graph propagation network.
\newblock \emph{IEEE Transactions on Knowledge and Data Engineering} {\bfseries
  35}\penalty0 (1):\penalty0 521--532.

\bibitem[Jin et~al.(2023)Jin, Zhang, Zhang, Meng, Zhang, Zhu, and
  Han]{jin2023patton}
Jin , B., Zhang , W., Zhang , Y., Meng , Y., Zhang , X., Zhu , Q., \& Han , J.
  (2023)
\newblock Patton: Language model pretraining on text-rich networks.
\newblock \emph{arXiv preprint arXiv:2305.12268}

\bibitem[Ju et~al.(2024)Ju, Fang, Gu, Liu, Long, Qiao, Qin, Shen, Sun, Xiao,
  et~al.]{ju2024comprehensive}
Ju~, W., Fang , Z., Gu~, Y., Liu , Z., Long , Q., Qiao , Z., Qin , Y., Shen ,
  J., Sun , F., Xiao , Z., \& others  (2024)
\newblock A comprehensive survey on deep graph representation learning.
\newblock \emph{Neural Networks} {\bfseries 173}:\penalty0 106207.

\bibitem[Khrabrov et~al.(2024)Khrabrov, Ber, Tsypin, Ushenin, Rumiantsev,
  Telepov, Protasov, Shenbin, Alekseev, Shirokikh, et~al.]{khrabrov20242}
Khrabrov , K., Ber , A., Tsypin , A., Ushenin , K., Rumiantsev , E., Telepov ,
  A., Protasov , D., Shenbin , I., Alekseev , A., Shirokikh , M., \& others
  (2024)
\newblock $\nabla$ 2 dft: A universal quantum chemistry dataset of drug-like
  molecules and a benchmark for neural network potentials.
\newblock \emph{SESAR Innovation Days} {\bfseries 2023}:\penalty0 82--83.

\bibitem[Kipf and Welling(2016)]{kipf2016semi}
Kipf , T.~N. \& Welling , M. (2016)
\newblock Semi-supervised classification with graph convolutional networks.
\newblock \emph{arXiv preprint arXiv:1609.02907}

\bibitem[Li et~al.(2024)Li, Wang, Zhu, Chen, Jiang, Cai, Chan, and
  Li]{li2024glbench}
Li~, Y., Wang , P., Zhu , X., Chen , A., Jiang , H., Cai , D., Chan , V.~W., \&
  Li~, J. (2024)
\newblock Glbench: A comprehensive benchmark for graph with large language
  models.
\newblock \emph{Advances in Neural Information Processing Systems} {\bfseries
  37}:\penalty0 42349--42368.

\bibitem[Liu et~al.(2013)Liu, Zhang, and Guo]{liu2013full}
Liu , X., Zhang , J., \& Guo , C. (2013)
\newblock Full-text citation analysis: A new method to enhance scholarly
  networks.
\newblock \emph{Journal of the American Society for Information Science and
  Technology} {\bfseries 64}\penalty0 (9):\penalty0 1852--1863.

\bibitem[Long et~al.(2020)Long, Jin, Song, Li, and Lin]{long2020graph}
Long , Q., Jin , Y., Song , G., Li~, Y., \& Lin , W. (2020)
\newblock Graph structural-topic neural network. In
\newblock \emph{Proceedings of the 26th ACM SIGKDD international conference on
  knowledge discovery \& data mining}
\newblock pages 1065--1073.

\bibitem[Long et~al.(2021)Long, Xu, Fang, and Song]{long2021hgk}
Long , Q., Xu~, L., Fang , Z., \& Song , G. (2021)
\newblock Hgk-gnn: Heterogeneous graph kernel based graph neural networks. In
\newblock \emph{Proceedings of the 27th ACM SIGKDD conference on knowledge
  discovery \& data mining}
\newblock pages 1129--1138.

\bibitem[Luo et~al.(2025)Luo, Zhang, Yuan, Zhao, Yang, Gu, Wu, Chen, Qiao,
  et~al.]{luo2025large}
Luo , J., Zhang , W., Yuan , Y., Zhao , Y., Yang , J., Gu~, Y., Wu~, B., Chen ,
  B., Qiao , Z., \& others  (2025)
\newblock Large language model agent: A survey on methodology, applications and
  challenges.
\newblock \emph{arXiv preprint arXiv:2503.21460}

\bibitem[Lv et~al.(2021)Lv, Ding, Liu, Chen, Feng, He, Zhou, Jiang, Dong, and
  Tang]{lv2021we}
Lv~, Q., Ding , M., Liu , Q., Chen , Y., Feng , W., He~, S., Zhou , C., Jiang ,
  J., Dong , Y., \& Tang , J. (2021)
\newblock Are we really making much progress? revisiting, benchmarking and
  refining heterogeneous graph neural networks. In
\newblock \emph{Proceedings of the 27th ACM SIGKDD conference on knowledge
  discovery \& data mining}
\newblock pages 1150--1160.

\bibitem[McCallum et~al.(2000)McCallum, Nigam, Rennie, and
  Seymore]{mccallum2000automating}
McCallum , A.~K., Nigam , K., Rennie , J., \& Seymore , K. (2000)
\newblock Automating the construction of internet portals with machine
  learning.
\newblock \emph{Information Retrieval} {\bfseries 3}:\penalty0 127--163.

\bibitem[Myers et~al.(2014)Myers, Sharma, Gupta, and Lin]{myers2014information}
Myers , S.~A., Sharma , A., Gupta , P., \& Lin , J. (2014)
\newblock Information network or social network? the structure of the twitter
  follow graph. In
\newblock \emph{Proceedings of the 23rd international conference on world wide
  web}
\newblock pages 493--498.

\bibitem[Schlichtkrull et~al.(2018)Schlichtkrull, Kipf, Bloem, Van Den~Berg,
  Titov, and Welling]{schlichtkrull2018modeling}
Schlichtkrull , M., Kipf , T.~N., Bloem , P., Van Den~Berg , R., Titov , I., \&
  Welling , M. (2018)
\newblock Modeling relational data with graph convolutional networks. In
\newblock \emph{The semantic web: 15th international conference, ESWC 2018,
  Heraklion, Crete, Greece, June 3--7, 2018, proceedings 15}
\newblock pages 593--607. Springer.

\bibitem[Sen et~al.(2008)Sen, Namata, Bilgic, Getoor, Galligher, and
  Eliassi-Rad]{sen2008collective}
Sen , P., Namata , G., Bilgic , M., Getoor , L., Galligher , B., \& Eliassi-Rad
  , T. (2008)
\newblock Collective classification in network data.
\newblock \emph{AI magazine} {\bfseries 29}\penalty0 (3):\penalty0 93--93.

\bibitem[Shi et~al.(2016)Shi, Li, Zhang, Sun, and Yu]{shi2016survey}
Shi , C., Li~, Y., Zhang , J., Sun , Y., \& Yu~, P.~S. (2016)
\newblock A survey of heterogeneous information network analysis.
\newblock \emph{IEEE Transactions on Knowledge and Data Engineering} {\bfseries
  29}\penalty0 (1):\penalty0 17--37.

\bibitem[Sinha et~al.(2015)Sinha, Shen, Song, Ma, Eide, Hsu, and
  Wang]{sinha2015overview}
Sinha , A., Shen , Z., Song , Y., Ma~, H., Eide , D., Hsu , B.-J., \& Wang , K.
  (2015)
\newblock An overview of microsoft academic service (mas) and applications. In
\newblock \emph{Proceedings of the 24th international conference on world wide
  web}
\newblock pages 243--246.

\bibitem[Tang et~al.(2008)Tang, Zhang, Yao, Li, Zhang, and
  Su]{tang2008arnetminer}
Tang , J., Zhang , J., Yao , L., Li~, J., Zhang , L., \& Su~, Z. (2008)
\newblock Arnetminer: extraction and mining of academic social networks. In
\newblock \emph{Proceedings of the 14th ACM SIGKDD international conference on
  Knowledge discovery and data mining}
\newblock pages 990--998.

\bibitem[Tang et~al.(2024)Tang, Yang, Wei, Shi, Su, Cheng, Yin, and
  Huang]{tang2024graphgpt}
Tang , J., Yang , Y., Wei , W., Shi , L., Su~, L., Cheng , S., Yin , D., \&
  Huang , C. (2024)
\newblock Graphgpt: Graph instruction tuning for large language models. In
\newblock \emph{Proceedings of the 47th International ACM SIGIR Conference on
  Research and Development in Information Retrieval}
\newblock pages 491--500.

\bibitem[Vashishth et~al.(2019)Vashishth, Sanyal, Nitin, and
  Talukdar]{vashishth2019composition}
Vashishth , S., Sanyal , S., Nitin , V., \& Talukdar , P. (2019)
\newblock Composition-based multi-relational graph convolutional networks.
\newblock \emph{arXiv preprint arXiv:1911.03082}

\bibitem[Velickovic et~al.(2017)Velickovic, Cucurull, Casanova, Romero, Lio,
  Bengio, et~al.]{velickovic2017graph}
Velickovic , P., Cucurull , G., Casanova , A., Romero , A., Lio , P., Bengio ,
  Y., \& others  (2017)
\newblock Graph attention networks.
\newblock \emph{stat} {\bfseries 1050}\penalty0 (20):\penalty0 10--48550.

\bibitem[Wang et~al.(2020)Wang, Shen, Huang, Wu, Dong, and
  Kanakia]{wang2020microsoft}
Wang , K., Shen , Z., Huang , C., Wu~, C.-H., Dong , Y., \& Kanakia , A. (2020)
\newblock Microsoft academic graph: When experts are not enough.
\newblock \emph{Quantitative Science Studies} {\bfseries 1}\penalty0
  (1):\penalty0 396--413.

\bibitem[Wang et~al.(2019)Wang, Ji, Shi, Wang, Ye, Cui, and
  Yu]{wang2019heterogeneous}
Wang , X., Ji~, H., Shi , C., Wang , B., Ye~, Y., Cui , P., \& Yu~, P.~S.
  (2019)
\newblock Heterogeneous graph attention network. In
\newblock \emph{The world wide web conference}
\newblock pages 2022--2032.

\bibitem[Wang et~al.(2022)Wang, Wang, Cao, and
  Barati~Farimani]{wang2022molecular}
Wang , Y., Wang , J., Cao , Z., \& Barati~Farimani , A. (2022)
\newblock Molecular contrastive learning of representations via graph neural
  networks.
\newblock \emph{Nature Machine Intelligence} {\bfseries 4}\penalty0
  (3):\penalty0 279--287.

\bibitem[White(2020)]{white2020pubmed}
White , J. (2020)
\newblock Pubmed 2.0.
\newblock \emph{Medical reference services quarterly} {\bfseries 39}\penalty0
  (4):\penalty0 382--387.

\bibitem[Yan et~al.(2023)Yan, Li, Long, Yan, Zhao, Zhuang, Yin, Zhang, Han,
  Sun, et~al.]{yan2023comprehensive}
Yan , H., Li~, C., Long , R., Yan , C., Zhao , J., Zhuang , W., Yin , J., Zhang
  , P., Han , W., Sun , H., \& others  (2023)
\newblock A comprehensive study on text-attributed graphs: Benchmarking and
  rethinking.
\newblock \emph{Advances in Neural Information Processing Systems} {\bfseries
  36}:\penalty0 17238--17264.

\bibitem[Yan et~al.(2024)Yan, Zhang, Fang, and Long]{yan2024inductive}
Yan , Y., Zhang , P., Fang , Z., \& Long , Q. (2024)
\newblock Inductive graph alignment prompt: bridging the gap between graph
  pre-training and inductive fine-tuning from spectral perspective. In
\newblock \emph{Proceedings of the ACM Web Conference 2024}
\newblock pages 4328--4339.

\bibitem[Yang et~al.(2021)Yang, Liu, Xiao, Li, Lian, Agrawal, Singh, Sun, and
  Xie]{yang2021graphformers}
Yang , J., Liu , Z., Xiao , S., Li~, C., Lian , D., Agrawal , S., Singh , A.,
  Sun , G., \& Xie , X. (2021)
\newblock Graphformers: Gnn-nested transformers for representation learning on
  textual graph.
\newblock \emph{Advances in Neural Information Processing Systems} {\bfseries
  34}:\penalty0 28798--28810.

\bibitem[Ye et~al.(2024)Ye, Ren, Wang, Wan, Razzak, Hoex, Wang, Xie, and
  Zhang]{ye2024construction}
Ye~, Y., Ren , J., Wang , S., Wan , Y., Razzak , I., Hoex , B., Wang , H., Xie
  , T., \& Zhang , W. (2024)
\newblock Construction and application of materials knowledge graph in
  multidisciplinary materials science via large language model.
\newblock \emph{Advances in Neural Information Processing Systems} {\bfseries
  37}:\penalty0 56878--56897.

\bibitem[Yu et~al.(2020)Yu, Shen, Li, and Lerer]{yu2020scalable}
Yu~, L., Shen , J., Li~, J., \& Lerer , A. (2020)
\newblock Scalable graph neural networks for heterogeneous graphs.
\newblock \emph{arXiv preprint arXiv:2011.09679}

\bibitem[Zhang et~al.(2024{\natexlab{a}})Zhang, Sun, Wang, Fan, Mo, Xu, Liu,
  Yang, and Shi]{zhang2024graphtranslator}
Zhang , M., Sun , M., Wang , P., Fan , S., Mo~, Y., Xu~, X., Liu , H., Yang ,
  C., \& Shi , C. (2024.
\newblock {\natexlab{a}}) Graphtranslator: Aligning graph model to large
  language model for open-ended tasks. In
\newblock \emph{Proceedings of the ACM Web Conference 2024}
\newblock pages 1003--1014.

\bibitem[Zhang et~al.(2024{\natexlab{b}})Zhang, Chen, Ma, Fang, and
  King]{zhang2024influential}
Zhang , X., Chen , Y., Ma~, C., Fang , Y., \& King , I. (2024.
\newblock {\natexlab{b}}) Influential exemplar replay for incremental learning
  in recommender systems. In
\newblock \emph{Proceedings of the AAAI Conference on Artificial Intelligence}
\newblock \emph{38}, pp. \penalty0 9368--9376.

\bibitem[Zhao et~al.(2022)Zhao, Qu, Li, Yan, Liu, Li, Xie, and
  Tang]{zhao2022learning}
Zhao , J., Qu~, M., Li~, C., Yan , H., Liu , Q., Li~, R., Xie , X., \& Tang ,
  J. (2022)
\newblock Learning on large-scale text-attributed graphs via variational
  inference.
\newblock \emph{arXiv preprint arXiv:2210.14709}

\end{thebibliography}
\normalsize
\newpage

\newpage
\appendix

\section{CITE Overview and Construction}\label{appendixA}

\subsection{Node \& Edge Distribution and Diversity}

\begin{table}[H]
\caption{Statistics of node and edges in CITE}
 \resizebox{\linewidth}{!}{
\begin{tabular}{ccccccc}
\toprule
{\textbf{Node Type}} & {\textbf{Numbers}} & {\textbf{Percentage}} & {\textbf{Edge Type}} & {\textbf{Numbers}} & {\textbf{Percentage}} & {\textbf{Source Coverage}} \\
\midrule
Paper  & {127690} & 29.13\% & Paper-Paper   & 335118 & 27.46\% & 96.5\% \\

Author & 221097    & 50.44\% & Paper-Author& 506327    & 41.49\% &  94.7\% \\
Keywords    & 86964& 19.84\% & Paper-Keywords& 253142    & 20.74\% &  39.7\% \\
Journal& 2553 & 0.53\%  & Paper-Journal & 125786    & 10.45\% & 98.5\%  \\ 
\bottomrule
\end{tabular}
}\label{statistics}
\end{table}
Table\ref{statistics} shows the statistics of nodes and edges in CITE. It outlines the quantities and proportions of various node types and edge types, along with their respective coverage, where Source Coverage represents the extent to which each type of edge covers the paper nodes. 
Despite the relatively small proportion of journal nodes within the entire network, the paper-journal edges cover 98\% of the paper nodes. Additionally, the majority of edges exhibit high coverage of paper nodes, suggesting that the dataset is both comprehensive and well-constructed. Specifically, the high coverage of paper nodes by different types of edges indicates that the paper nodes possess a rich and comprehensive set of heterogeneous information, resulting in a densely interconnected network. This reflects structural integrity of CITE, thereby providing a solid foundation for subsequent analyses and modeling tasks.
While authors account for 50\% of nodes reflecting domain authorship dynamics, this majority isn't disproportionately amplified in edge formation - author-paper relationships comprise 41\% of edges. 
Label distributions span 85 major subfields with no single category exceeding 25\% of total papers, unlike existing datasets where >25\% of papers may cluster in single subfields (e.g., CV or LG). This balance mitigates training biases in tasks like classification or recommendation, while the integration of temporal (1922–2022) and textual attributes supports fine-grained analyses of emerging trends. 
\begin{table}[H]
\centering
\caption{Missing Value Rates(\%) for Key Metadata}
\label{detail}
 \resizebox{0.7\linewidth}{!}{
\begin{tabular}{cccc}
\toprule
\textbf{Metadata}  & \textbf{Missing Value Rate}           & \textbf{Metadata}& \textbf{Missing Value Rate}\\ \midrule
id     & 0  &
doi    & 0   \\
title  & 0   &
year   & 0.001 \\
authors    & 0 &
first author &  0.001\\
publish\_date    & 0.005        &
volume &   2.077 \\
page\_start &1.458&
page\_end &  3.773\\
print\_issn &  2.488   &
abstract    & 3.463 \\
journal& 1.594  &
cited\_times& 0 \\
reference\_times & 0  &
organization&  0 \\
citations   & 0    &
references  & 0   \\
authors\_split &  0 \\\bottomrule
\end{tabular}
}
\end{table}
\subsection{Composition of CITE}
\textbf{Node Construction and Definition}:
CITE comprises four core entity types: Paper, Author, Journal, and Keywords. Each node type is assigned a unique identifier through a domain-specific alignment pipeline, ensuring semantic consistency and resolving potential conflicts.
The attribute interpretation and alignment rules are defined as follows:
\begin{longtable}{L{3cm} L{10cm}} 
\toprule
\textbf{Node Type} & \textbf{Description} \\
\hline
\endfirsthead
\hline
\endfoot
\hline
Paper & Represent the basic content of the article, including the title and abstract, which are extracted from the metadata. The labels are defined based on the SCI subject classification of the journal to which the paper belongs. If the journal is not indexed in SCI, the paper is labeled as 'uncategorized'. Unique Identifier is generated by: paper\_ID = "j\_" + MD5(title,DOI)[:8], we also have a simplified version, which is a numeric type from 1 to 127689.\\
\hline
Author & Represent the authors of articles, have a high probability of name duplication. We distinguish authors with the same name using ORCID. For author nodes with the same name and no ORCID, if they belong to the same institution, we consider them to be the same person. In addition, the name order is uniformly replaced with first name + last name. We also have two Unique Identifiers, MD5 and numeric, the latter is from 0 to 220196.\\
\hline
Journal & Represent the journals or conferences where papers are published. Through metadata extraction, we also extract the ISSN of the corresponding data and cross-reference it with the ISSN International Center’s records to convert the journal names to their official and formal titles. (e.g., “J. Catal.” → “Journal of Catalysis”). The Unique Identifier of the journal nodes is numeric, from 0 to 2552.\\
\hline
Keywords & Represent the keywords contained in the paper abstract, which typically reflect the core research issues and key technologies involved in the study, so we do not conduct other manipulation. The Unique Identifier of the Keyword nodes is numeric, from 0 to 86963.\\
\end{longtable}

\textbf{Edge Generation Rules and Types}: The edges in our graph are designed to capture both explicit relationships (e.g., citations, authorship) and implicit semantic connections (e.g., topical similarity). We define four types of edges: Paper-Paper, Paper-Author, Paper-Journal, Paper-Keywords.
\begin{longtable}{L{3cm} L{10cm}}
\toprule
\textbf{Edge Type} & \textbf{Description} \\
\hline
\endfirsthead
\hline
\endfoot
\hline
Citation (Paper-Paper) & Generated based on explicit citation records in Crossref metadata reference lists. Matching the target paper node by the generated Unique Identifier.\\
\hline
Authorship (Paper-Author)  & Derive from the author list in the metadata, with each author uniquely identified by a Unique Identifier to ensure no duplication or omission.\\
\hline
Publication  (Paper-Journal) & Associates journal nodes through the ISSN of each paper in the metadata.\\
\hline
Topical Relevance (Paper-Keywords) & Generated by extracting keywords from the metadata, followed by data cleaning to correct anomaly. \\
\end{longtable}

\subsection{Broader Impact}
\textbf{Positive Social Impact}\newline
This dataset provides a cross-disciplinary knowledge graph infrastructure for accelerating clean energy technology innovation by modeling a heterogeneous citation network of 127,690 photocatalytic materials. First, the coupled analysis of journals and keywords can reveal cross-technology pathways (e.g., the migration potential of photovoltaic materials in catalysis) that have been neglected by mainstream research, reducing the trial-and-error cost of new material discovery. Second, the dynamic characterization of author collaboration networks can help identify the original contributions of nascent research teams, counter the citation monopoly of academic authorities, and promote research equity. In addition, the mining of high-potential papers in long-tail journals can challenge the traditional impact factor evaluation system and provide a decentralized basis for resource allocation. These characteristics make the dataset an instrumental asset for advancing the achievement of the Sustainable Development Goals (SDGs).\newline
\textbf{Negative social impact}\newline
Despite its scientific value, the application of the dataset may pose potential risks. First, the lack of keyword coverage and English-language journal preference may lead to algorithms that reinforce the dominant narrative and suppress innovation in localized materials from non-English-speaking academic communities. Second, the strong correlation between journal tiers and citation counts may be abused as a criterion for judging the quality of papers, exacerbating “top journal” oriented academic alienation and further marginalizing practical research. More insidiously, topological analysis of citation networks may be used by commercial entities to identify high-value technology paths, forming a “thesis-patent” closed loop through patent barriers, and ultimately hindering open-source sharing of technology. These risks require users to strictly follow the bias correction protocols in the Data Deficiencies Statement.
\subsection{Workflow for Constructing CITE}
The following algorithm shows the processing from a json file to csv files of CITE.

\begin{algorithm}
\caption{Preprocess workflow}
\begin{algorithmic}[1]

\STATE \textbf{Input:} Json file $F$ of metadata, including $P_F, A_F, J_F, K_F, I_F$, which denotes set of Paper(Title \& Abstract), Author(Name, Orcid), Journal(Journal Name), Keywords, ISSN
\STATE \textbf{Output:} CITE-$G(V,E,T)$, where $V$ denotes node set, $E$ denotes edge set, $T$ denotes text set
\STATE Assign Unique Identifier $p_i, a_i$ for $p_F\in P_F$, $a_F\in A_F$
\STATE Construct Paper set $V_P \in V$, Author set $V_A \in V$
\FOR{item $i$ in $F$}
    \STATE Construct CSV column $C_i$
\ENDFOR
\FOR{item in Journal column $j_C \in J_C$}
    \STATE Standardize $j_C$ with $i_C$, construct journal set $V_J\in V$
    \STATE Assign Unique Identifier $j_i$
\ENDFOR
\FOR{item in Keywords column $k_C \in K_C$}
    \STATE Construct keywords set $V_K\in V$
    \STATE Assign Unique Identifier $k_i$
\ENDFOR
\FOR{row $i$ in CSV}
    \IF{$j_i$} \STATE Construct paper-journal edge set $E_J \in E$, text set $T_J \in T$
    \STATE Assign label $L_i$ 
    \ELSIF{$k_i$} \STATE Construct paper-keywords edge set $E_K \in E$, text set $T_K \in T$
    \ELSIF{$a_i$} \STATE Construct paper-author edge set $E_A \in E$, text set $T_A \in T$
    \ELSIF{$p_i$(citation/reference)} \STATE Construct paper-paper edge set $E_P \in E$, text set $T_P \in T$
    \ENDIF
\ENDFOR

\end{algorithmic}
\end{algorithm}

\subsection{Metadata Example and Explanation}
The following subsection provides a sample JSON file and a detailed explanation of each key corresponding to the metadata discussed in Section 3.2.
\lstdefinestyle{jsonstyle}{
    basicstyle=\ttfamily\small,        
    breaklines=true,               
    backgroundcolor=\color{gray!5},    
    moredelim=[s][\color{blue}]{/*}{*/}, 
    showstringspaces=false             
}

\begin{tcolorbox}[colback=gray!5!white, colframe=gray!80!black, title=Example of json file part1.]
\begin{lstlisting}[style=jsonstyle]
{
    "id": "j_b157513a0109502b2e282233f6c7359b",
    "doi": "10.1039/c3sc52323c",
    "year": "2014",
    "zh_title": "BODIPY triads triplet photosensitizers enhanced with intramolecular resonance energy transfer (RET): broadband visible light absorption and application in photooxidation",
    "authors": "Song Guo;Lihua Ma;Jianzhang Zhao;...,
    "firstauthor": "Song Guo",
}
\end{lstlisting}
\end{tcolorbox}

\lstdefinestyle{jsonstyle}{
    basicstyle=\ttfamily\small,        
    breaklines=true,       
    backgroundcolor=\color{gray!5},   
    moredelim=[s][\color{blue}]{/*}{*/}, 
    showstringspaces=false            
}

\begin{tcolorbox}[colback=gray!5!white, colframe=gray!80!black, title=Example of json file part2.]
\begin{lstlisting}[style=jsonstyle]
{
    
    "authors_split": [
        {
            "person_id": "640bdfe1fb47722feb2617a8e78d424e",
            "name": "Jianzhang Zhao",
            "email": "zhaojzh@dlut.edu.cn",
            "org_name": "Dalian Polytechnic University",
            "org_id": "cf1e4a35f2dc4ed013ed461695745894",
            "author_num": 3,"nationality": "China",
            "country": "China","province": "Liaoning.","city": "Dalian."
        },...
    ],
    "hasfulltext": true,"publish_date": "2014-01-01",
    "volume": "5","issue": "2","page_start": "489",
    "page_end": "500","print_issn": "2041-6520;2041-6539",
    "abstract": "Resonance energy transfer (RET) ...,
    "field_name": "Chemistry","journal": "Chem. Sci.",
    "source": ["Crossref","corpus","unpaywall","openalex","scimag"],
    "cited_times": "93","reference_times": "100","is_oa": false,
    "tags": [],"oa_status": "closed","journal_tags": [],
    "organization": [
        {"org_name": "Tianjin Polytechnic University","org_id": "00490b72bc0ea69a22833962cb59e2fe"},...
    ],
    "citation": [
        {
            "id": "j_bfa6110db9f53269a7f251fdd6a0fdf0",
            "zh_title": "A mono(carboxy)porphyrin-...",
            "authors": "Ganesh D. Sharma;...,
            "journal": "J. Mater. Chem. C","year": "2015"
        },...
    ],
    "references": [
        {
            "id": "j_25c197227b70da7c79685f22632f506a",
            "zh_title": "Light-Harvesting Fullerene Dyads as ...,
            "authors": "Wanhua Wu;Jianzhang Zhao;Jifu Sun;Song Guo",
            "journal": "JOURNAL OF ORGANIC CHEMISTRY",
            "year": "2012"
        },...
    ]
}
\end{lstlisting}
\end{tcolorbox}

\begin{table}[H]
\caption{Csv columns and detailed information}
\label{detail}
 \resizebox{\linewidth}{!}{
\begin{tabular}{@{}llll@{}}
\toprule
\textbf{Field}  & \textbf{Description}           & \textbf{Example}\\ \midrule
id     & unique identifier of paper & j\_000033981dfae15198d714aad50b7e51  \\
doi    & Paper DOI number  & 10.1016/j.jhazmat.2011.07.096   \\
title  & paper title    & Fabrication of WO3/Cu2O composite films and...   \\
year   & Year of publication         & 2011 \\
authors    & List of authors in order  & Shouqiang Wei;Yuyan Ma;Yuye Chen;Long Liu.... \\
first author & The first author of the paper & Shouqiang Wei\\
publish\_date    & Specific date of publication        & 2011/10/1         \\
volume &  a collection of all issues& 194 \\
page\_start &start page of paper&243\\
page\_end & end page of paper&249\\
print\_issn & Journal Print ISSN Number    & 0920-5861   \\
abstract    & Abstract text & Spherical core-shell Ag/ZnO nanocomposites ... \\
journal& Standardized name of the journal     & Catalysis Today  \\
cited\_times& Number of citations      & 18  \\
reference\_times & Number of citations to other papers & 50  \\
organization& Authors' affiliation    & \{'org\_name': 'Serbian....', 'org\_id': 'e0fef...'\};... \\
citations   & List of papers that cite this article  & \{'id': 'j\_23d26fe01....', 'zh\_title': 'Facile preparation...    \\
references  & List of papers cited in this article  & \{'id': 'j\_795d75c0...', 'zh\_title': 'Ag/ZnO Hetero...   \\
authors\_split   & Standardized author list     & \{'person\_id': '87230cc3..', 'orcid': 'https://orcid... \\
fields & Label of research areas of their journals    & CHEMISTRY, APPLIED; ...  \\ \bottomrule
\end{tabular}
}
\end{table}

\section{Experiment and Evaluation}\label{appendixB}
\subsection{Hyperparameter Settings}
The following subsection provides detailed information about the models' hyperparameters, with each setting being the model's best configuration.
\begin{longtable}{L{3cm} L{10cm}}
\toprule
\textbf{Model} & \textbf{Hyperparameter} \\
\hline
\endfirsthead
\hline
\endfoot
\hline
GCN, GAT, GraphSAGE & num\_layers in [2, 3, 4], hidden\_dim in [64, 128, 256], dropout in [0.3, 0.5, 0.6]\\
\hline
RGCN & learning\_rate = 0.01, weight\_decay = 0.0001, dropout = 0.2
        seed = 0, in\_dim = 64,\newline hidden\_dim = 64, n\_bases = 40, fanout = 4
        num\_layers = 3, max\_epoch = 50, patience = 50, batch\_size = 128\\
\hline
CompGCN & num\_layers = 2, in\_dim = 32, hidden\_dim = 32, out\_dim = 32, max\_epoch = 500,
        patience = 100, comp\_fn = sub, validation = True, mini\_batch\_flag = True, batch\_size = 128, fanout = 4\\
        \hline
SimpleHGN & hidden\_dim = 256, num\_layers = 3, num\_heads = 8, feats\_drop\_rate = 0.2, slope = 0.05, edge\_dim = 64, seed = 0, max\_epoch = 500, patience = 100, lr = 0.001, weight\_decay = 5e-4, beta = 0.05, residual = True, mini\_batch\_flag = True, fanout = 5, batch\_size = 2048 \\
\hline
HPN&
        seed = 0, learning\_rate = 0.005, weight\_decay = 0.001, dropout = 0.6, k\_layer = 2, alpha = 0.1, edge\_drop = 0, hidden\_dim = 64, out\_dim = 16, max\_epoch = 200, patience = 100\\
\hline
HGT&
    seed = 0, learning\_rate = 0.001, weight\_decay = 0.0001, dropout = 0.4, batch\_size = 5120, patience = 40, hidden\_dim = 64, out\_dim = 16, num\_layers = 2, num\_heads = 8, num\_workers = 64, max\_epoch = 500, mini\_batch\_flag = True, fanout = 5\\
\hline
NARS&
    seed = 0, learning\_rate = 0.003, weight\_decay = 0.001, dropout = 0.7, hidden\_dim = 64, out\_dim = 16, num\_heads = 8, num\_hops = 2, max\_epoch = 200, mini\_batch\_flag = False, R = 2, patience = 100, input\_dropout = True, cpu\_preprocess = True, ff\_layer = 2\\
\hline
TAPE&
    num\_layers in [2, 3, 4], hidden\_dim in [64, 128, 256], dropout in [0.3, 0.5, 0.6]
\end{longtable}

\subsection{Execution Time and Resource Utilization}
\begin{figure}[H]
  \centering
  \begin{minipage}{0.45\textwidth} 
    Figure \ref{execution} illustrates detailed execution metrics of the models. The x-axis represents the logarithm of execution time, while the y-axis corresponds to the models' F1 scores. The size of each point reflects the number of parameters of the respective model. All models were executed on an NVIDIA A100 GPU with 80GB of memory. The homogeneous graph models have parameter scales of 8 million. Among the heterogeneous graph models, CompGCN has a parameter scale of 14 million, RGCN, HPN, HGT, NARS has a parameter scale of 28 million, and SimpleHGN has a parameter scale of 120 million. The LLM-centric model and LLM+Graph models have a parameter scale of 7 billion.
  \end{minipage}
  \hfill
  \begin{minipage}{0.5\textwidth} 
    \centering
    \includegraphics[width=\textwidth]{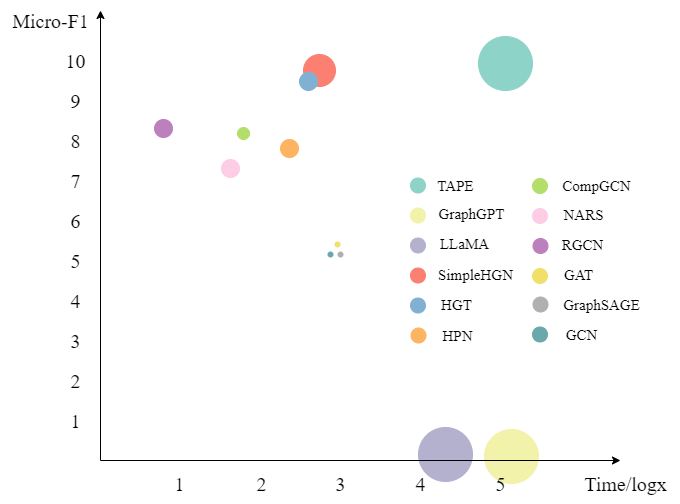}
    \caption{Execution Information}
    \label{execution}
  \end{minipage}
\end{figure}
\subsection{Sensitivity Evaluation of Experiments}
\begin{figure}[H]
    \centering
    \begin{subfigure}[b]{0.45\textwidth}
        \centering
        \includegraphics[width=\textwidth]{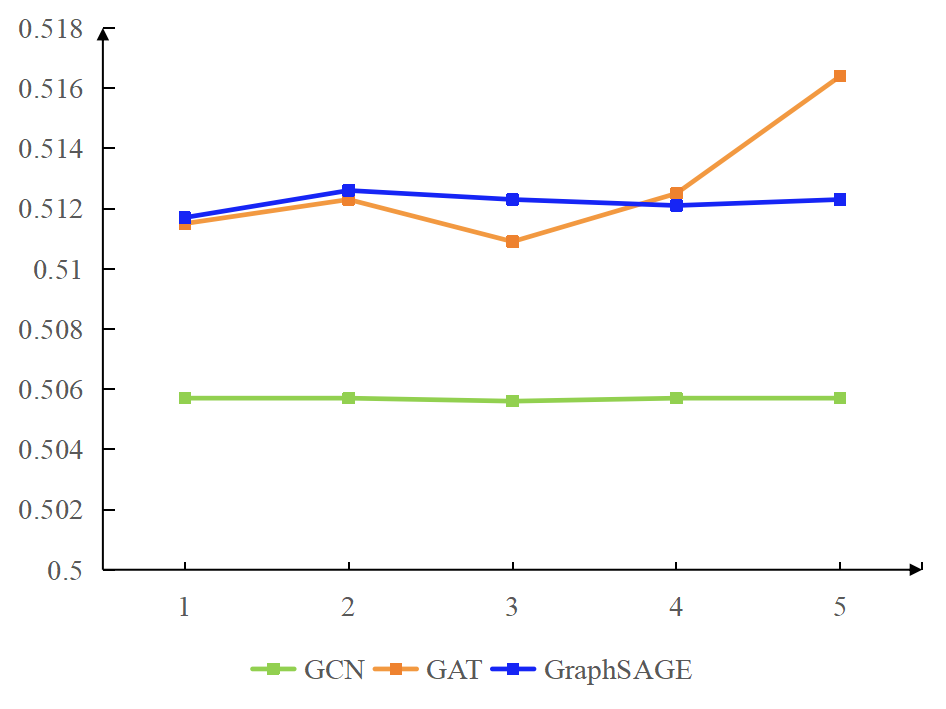}
        \caption{Micro-F1 of homogeneous graph models}
    \end{subfigure}
    \hspace{0.1cm}
    \begin{subfigure}[b]{0.45\textwidth}
        \centering
        \includegraphics[width=\textwidth]{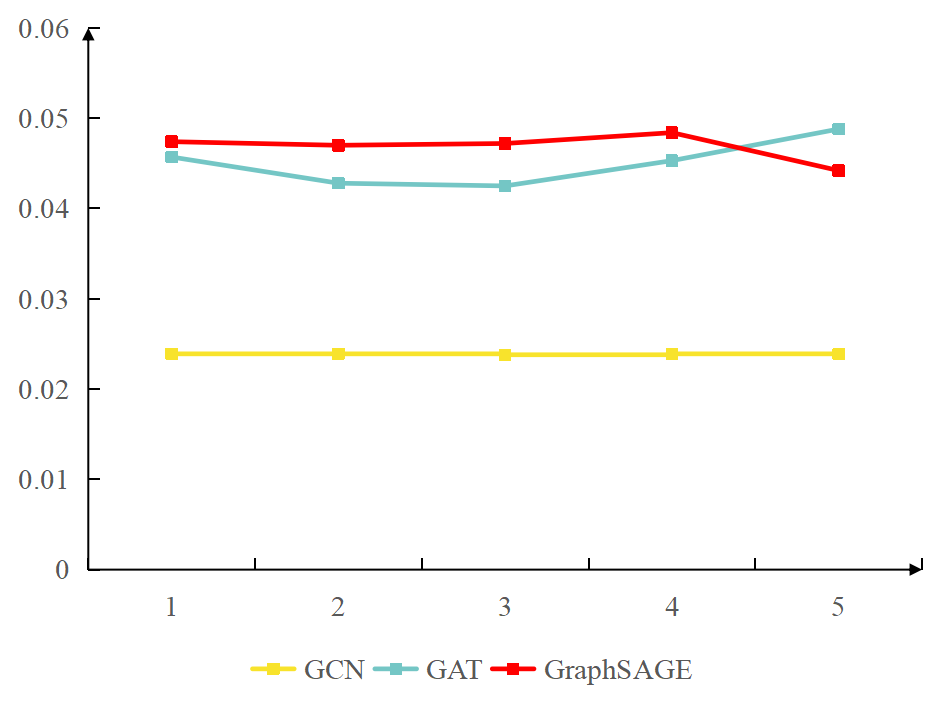}
        \caption{Macro-F1 of homogeneous graph models}
    \end{subfigure}
 \vspace{0.4cm}
    \begin{subfigure}[b]{0.45\textwidth}
        \centering
        \includegraphics[width=\textwidth]{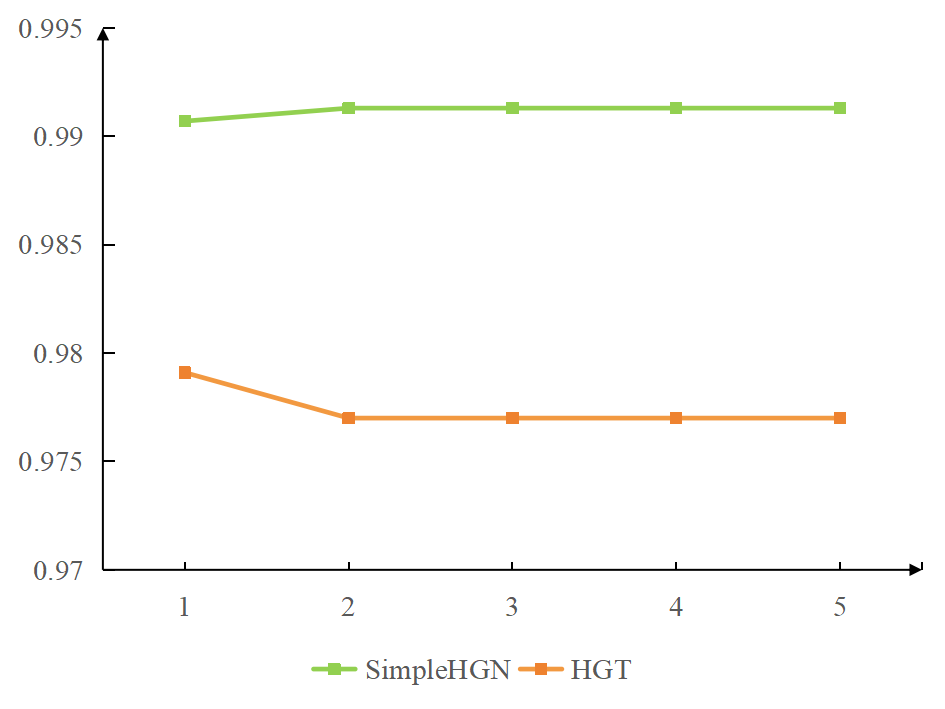}
        \caption{Micro-F1 of heterogeneous graph models}
    \end{subfigure}
    \hspace{0.1cm}
    \begin{subfigure}[b]{0.45\textwidth}
        \centering
        \includegraphics[width=\textwidth]{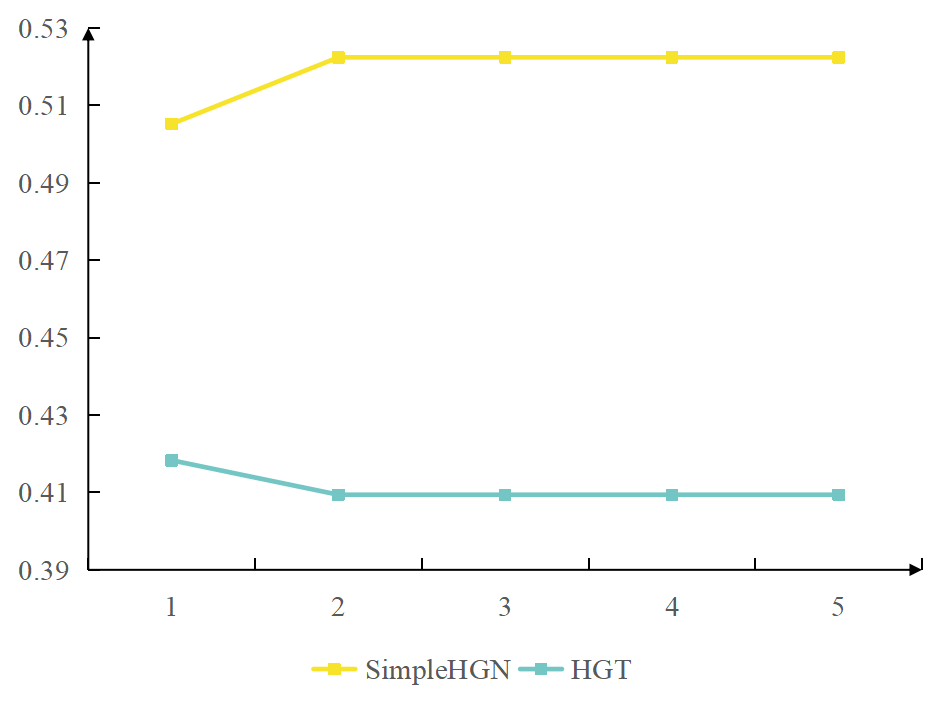}
        \caption{Macro-F1 of heterogeneous graph models}
    \end{subfigure}
 \vspace{0.4cm} 
    \begin{subfigure}[b]{0.45\textwidth}
        \centering
        \includegraphics[width=\textwidth]{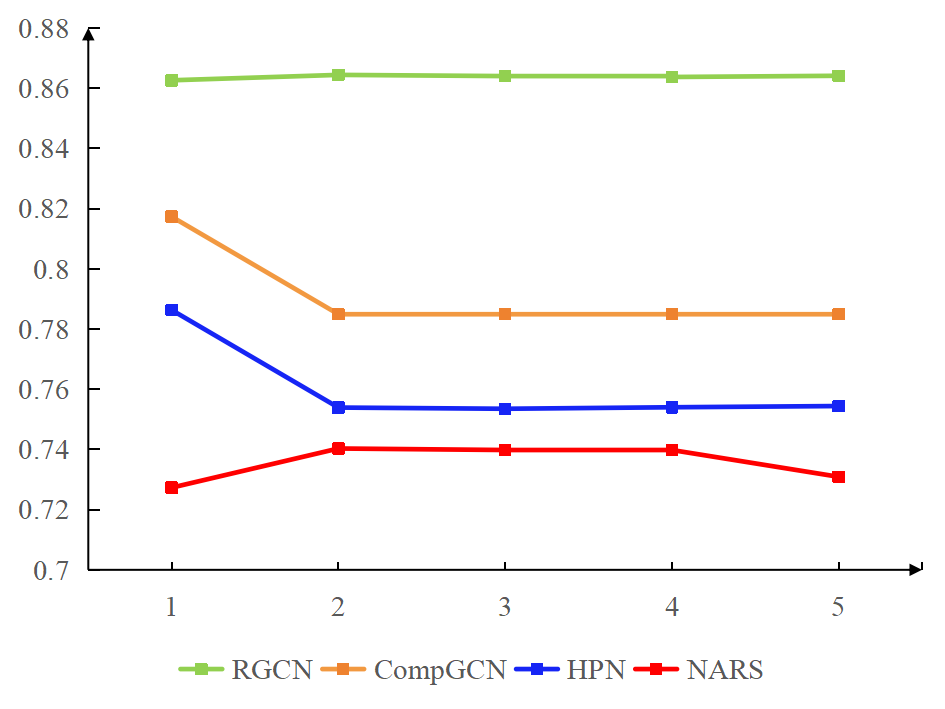}
        \caption{Micro-F1 of heterogeneous graph models}
    \end{subfigure}
    \hspace{0.1cm}
    \begin{subfigure}[b]{0.45\textwidth}
        \centering
        \includegraphics[width=\textwidth]{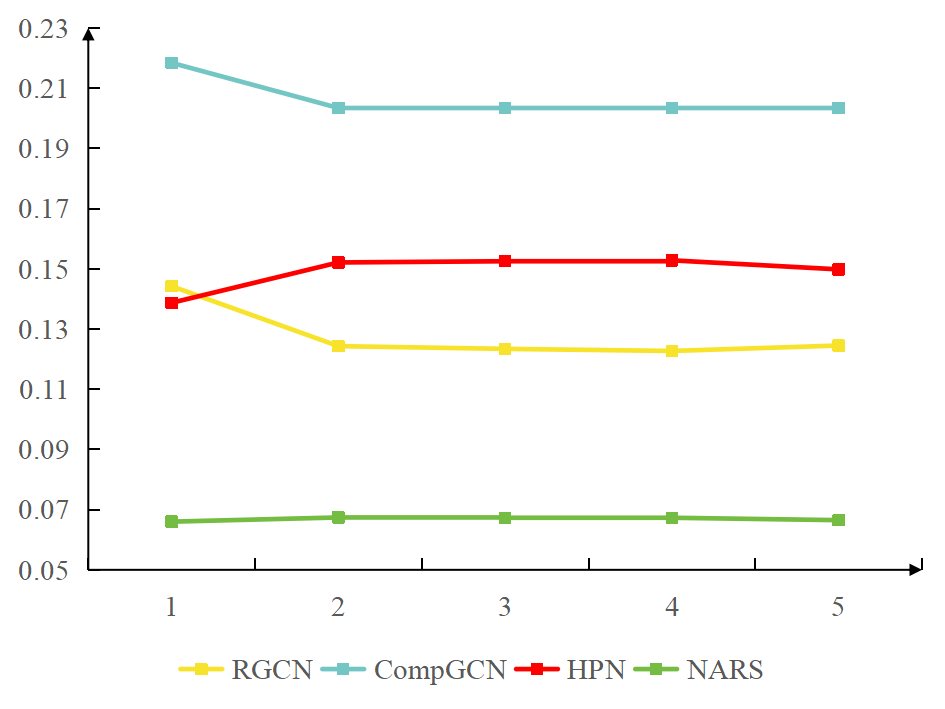}
        \caption{Macro-F1 of heterogeneous graph models}
    \end{subfigure}
    
    \vspace{0.4cm}
    
    \begin{subfigure}[b]{0.45\textwidth}
        \centering
        \includegraphics[width=\textwidth]{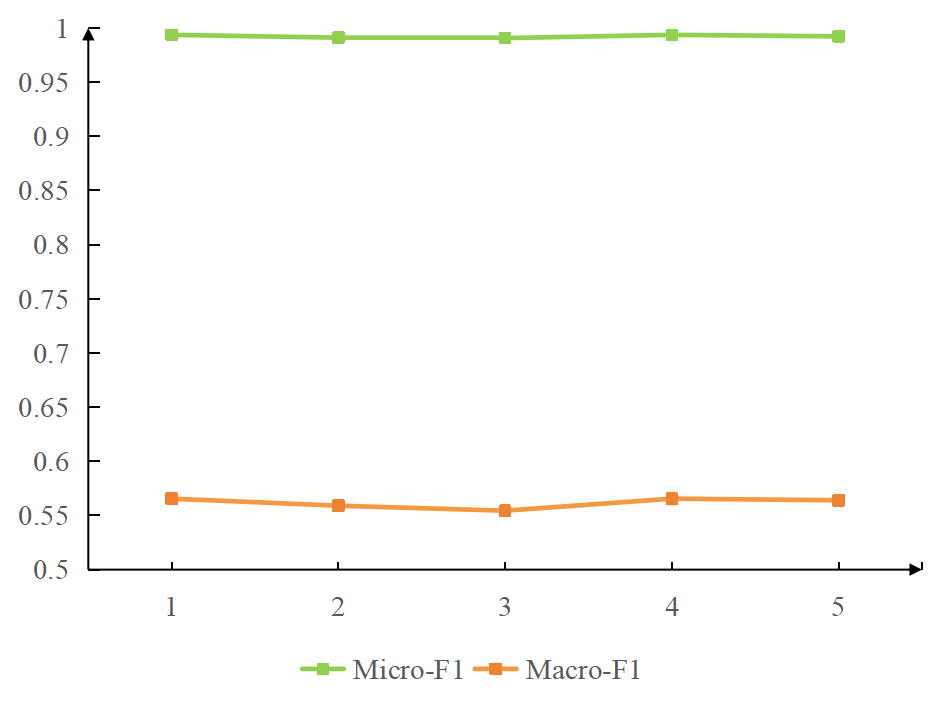}
        \caption{Micro-F1/Macro-F1 of TAPE}
    \end{subfigure}
    \hspace{0.1cm}
    \begin{subfigure}[b]{0.45\textwidth}
        \centering
        \includegraphics[width=\textwidth]{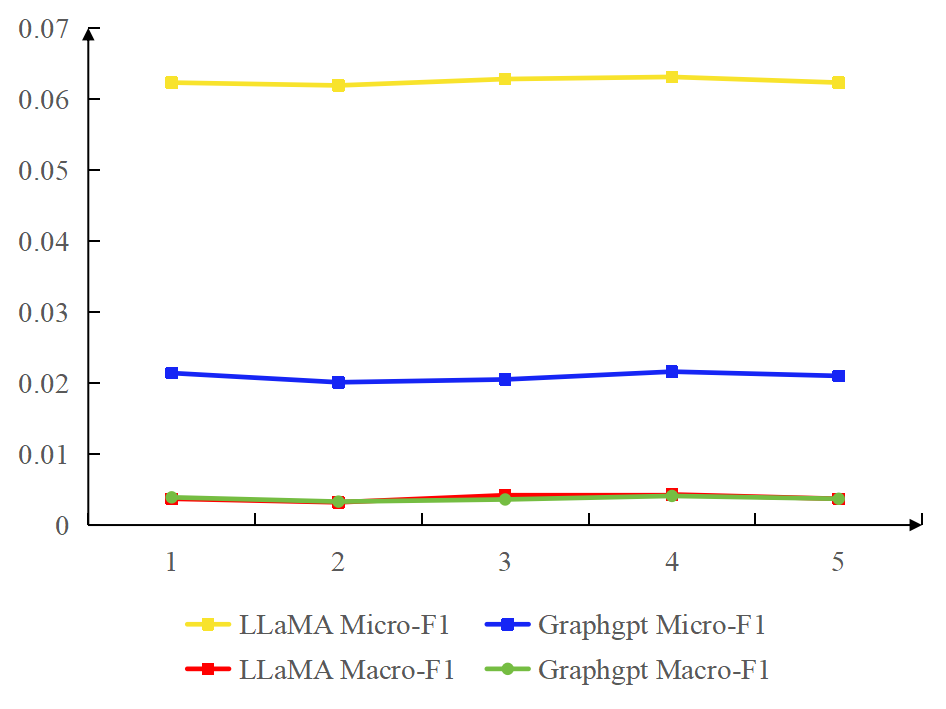}
        \caption{Micro-F1/Macro-F1 of LLaMA and GraphGPT}
    \end{subfigure}
    
    \caption{Performance of models in duplicate experiments}
    \label{B2}
\end{figure}

Based on  Figure\ref{B2} and Table\ref{6}, we discovered distinct sensitivity patterns across model architectures during multi-trial executions. For homoegenous graph models, we observe an inverse correlation between model performance and output stability - lower-performing models variants exhibit minimal result variations (2$\sigma \leq 0.0001$) across repeated trials, while the higher-performing counterparts demonstrate substantially greater fluctuations (2$\sigma > 0.001$). This pattern is reversed in heterogeneous graph models, where top-performing models maintain better consistency ($\sigma < 0.01$) compared to underperforming models that show heightened sensitivity ($\sigma > 0.01$). Notably, LLM-integrated systems display different characteristics, with nearly all models maintaining  $\sigma > 0.003$. The LLM-as-predictor paradigm achieves particular stability, outperforming all heterogeneous graph models. This suggests that LLMs introduce an inherent robustness to stochastic training dynamics, potentially attributable to their massive parameter space and pre-training regularization effects. Our experimental findings suggest future architectures could strategically combine LLM-integrated system's stability with heterogeneous graph models' data processing mechanisms.
\begin{table}[htbp]
\caption{Results and uncertainties of baseline model comparison (mean ± 2$\sigma$)}
\centering
\begin{tabular}{cccc}
\toprule
\textbf{Category}  & \textbf{Model Name} & \textbf{Micro-f1} & \textbf{Macro-f1} \\ \midrule
\multirow{3}{*}{\begin{tabular}[c]{@{}c@{}} Homogeneous Graph Model\end{tabular}}  
& GCN & 0.5057$\pm$0.0001& 0.0239$\pm$0.0001          \\
 & GAT                 & 0.5115 $\pm$0.0043           & 0.0457  $\pm$0.0051          \\
 & GraphSAGE           & 0.5117 $\pm$0.0007             & 0.0474$\pm$0.0031           \\ \midrule
\multirow{6}{*}{\begin{tabular}[c]{@{}c@{}}Heterogeneous Graph Model\end{tabular}} & RGCN                & 0.8626$\pm$0.0018            & 0.1442$\pm$0.0215            \\
 & CompGCN             & 0.8173$\pm$0.0324            & 0.2184$\pm$0.0150            \\
 & SimpleHGN           & \colorbox{yellow!60}{0.9907$\pm$0.0005}           & \colorbox{yellow!60}{0.5052$\pm$0.0172}            \\
   & HPN                 & 0.7863$\pm$0.0328            & 0.1387$\pm$0.0141           \\
  & HGT                 & \colorbox{cyan!60}{0.9791$\pm$0.0019}            & \colorbox{cyan!60}{0.4183$\pm$0.0089}           \\
 & NARS                & 0.7273 $\pm$0.0130           & 0.0660$\pm$0.0014           \\ \midrule
LLM-centric Model  & LLaMA  & 0.0614$\pm$0.0015    & 0.0037$\pm$0.0006      \\ \midrule
\multirow{2}{*}{LLM+Graph Model}  & Graphgpt (Vicuna 7B) & 0.0214$\pm$0.0013               & 0.0039$\pm$0.0006               \\
 & TAPE (LLaMA 7B)      & \colorbox{red!60}{0.9936$\pm$0.0030}            & \colorbox{red!60}{0.5655$\pm$0.0111}            \\ \bottomrule
\end{tabular}

\label{6}
\end{table}
\newpage

\section{Details of input and output}\label{appendixC}
\subsection{LLM prompts}
\begin{tcolorbox}[colback=gray!5!white, colframe=gray!80!black, title=GraphGPT prompts., enhanced, breakable]
\textbf{Question of Stage1:}\\
Given a sequence of graph tokens \verb|<graph>| that constitute a subgraph... \\
Title: \texttt{\{title\}}, \\
Abstract: \texttt{\{abstract\}}, \\
Journal: \texttt{\{journal\}}, \\
Authors: \texttt{\{authors\}}, \\
Keywords: \texttt{\{key\_words\}}.\\
\textbf{Question of Stage2:} \\
Given a citation graph: \verb|<graph>| where the 0th node is the target paper... \\
Title: \texttt{\{center\_title\}}, \\
Abstract: \texttt{\{center\_abstract\}}, \\
Journal: \texttt{\{center\_journal\}}, \\
Authors: \texttt{\{center\_authors\}}, \\
Keywords: \texttt{\{center\_key\_words\}}.\\
Question: Which SCI category does this paper belong to? ... \\
\end{tcolorbox}
\noindent

\begin{tcolorbox}[colback=gray!5!white, colframe=gray!80!black, title=LLaMA\&TAPE prompts., enhanced, breakable]
\textbf{Question:}\\
Title: \texttt{\{title\}}, \\
Abstract: \texttt{\{abstract\}}, \\
Journal: \texttt{\{journal\}}, \\
Authors: \texttt{\{authors\}}, \\
Keywords: \texttt{\{key\_words\}}. \\
Which of the following categories does this paper belong to:... \\
\end{tcolorbox}

In the ablation experiments conducted for Q3, the corresponding components were removed from the prompt while ensuring that the overall integrity and structure of the prompt remained intact.

\subsection{Responses of LLM}
\begin{tcolorbox}[colback=gray!5!white, colframe=gray!80!black, title=GraphGPT responses., enhanced, breakable]
\textbf{True one:}Based on the given citation, the paper belongs to the category "CHEMISTRY" \\
\textbf{False one:} Based on the given citation, the paper is likely to belong to the category "Nanoparticles of CdS and Indole"(not included in the subject list) \\
\textbf{Typica false response:} Based on the given citation, we obtain the following results: 1.ENGINEERING 2.MATERIALS SCIENCE 3.PHYSICS 4.CHEMISTRY...(repeat the subject category and response nothing else) \\
\end{tcolorbox}
The analysis of the LLM responses below reveals two prevalent error types encountered during node classification tasks: (1) outputs that fall outside the predefined label set, resulting in invalid predictions; and (2) instances where the model redundantly repeats the input query without producing any prediction. These errors underscore the limitations of LLMs in handling classification tasks involving a large number of categories.
\begin{tcolorbox}[colback=gray!5!white, colframe=gray!80!black, title=LLaMA\&TAPE responses., enhanced, breakable]
\textbf{True one:} The paper belongs to the category of CHEMISTRY. \\
\textbf{False one:} The paper belongs to the category of SCIENCE RESEARCH (not included in the subject list)\\
\textbf{Typica false response:} Title: \texttt{\{title\}},Abstract: \texttt{\{abstract\}}...(repeat the graph information and response nothing else)
\end{tcolorbox}

\subsection{Information of Models and Datasets}

The models evaluated in this paper is introduced as follows,
\begin{itemize}
    \item \textbf{GCN.}~\cite{kipf2016semi}Graph Convolutional Network (GCN) is a classical model that works by performing a linear approximation to spectral graph convolutions
    \item \textbf{GAT.}~\cite{velickovic2017graph}Graph Attention Network (GAT) introduces the attention mechanism to capture the importance of neighboring nodes when aggregating information from the graph.
    \item \textbf{GraphSAGE.}~\cite{hamilton2017inductive}GraphSAGE is a GNN model that focuses on inductive node classification, but can also be applied for transductive settings.
    \item \textbf{RGCN.}~\cite{schlichtkrull2018modeling} Relational Graph Convolutional Networks (R-GCNs) are neural networks designed to model multi-relational knowledge graphs by effectively aggregating information across diverse relation types. They improve tasks like link prediction and entity classification by capturing complex graph structures and relations
    \item \textbf{CompGCN.}~\cite{vashishth2019composition} CompGCN is a graph convolutional framework that jointly learns embeddings for nodes and relations in multi-relational graphs. It incorporates entity-relation composition operations from knowledge graph embedding techniques and generalizes existing multi-relational GCN methods, enabling scalable modeling with respect to the number of relations.
    \item \textbf{SimpleHGN.}~\cite{lv2021we} SimpleHGN is a graph neural network model designed for heterogeneous graphs. It simplifies the message-passing process between different types of nodes and relations to efficiently aggregate features, enabling effective modeling of diverse relational structures commonly found in heterogeneous graphs.
    \item \textbf{HPN.}~\cite{ji2021heterogeneous} HPN is a heterogeneous graph neural network that addresses semantic confusion in deep models by weighting local node semantics during aggregation and learning the importance of meta-paths for effective fusion, improving node embedding distinguishability.
    \item \textbf{HGT.}~\cite{hu2020heterogeneous} Heterogeneous Graph Transformer (HGT) is a graph neural network designed for large-scale heterogeneous graphs. It uses node- and edge-type specific parameters to model diverse relations via heterogeneous attention mechanisms. HGT employs a mini-batch sampling algorithm (HGSampling) for scalable training and demonstrates superior performance on large datasets like the Open Academic Graph.
    \item \textbf{NARS.}~\cite{yu2020scalable} NARS is a method for heterogeneous graphs that trains classifiers on neighbor-averaged features from randomly sampled relation subgraphs. It optimizes memory efficiency during training and inference and achieves state-of-the-art accuracy on multiple benchmarks, outperforming more complex GNN models.
    \item \textbf{LLaMA-7B.} LLaMA-7B is a large language model with 7 billion parameters designed for efficient natural language understanding and generation. It balances performance and computational resources, enabling effective use in various NLP tasks while maintaining scalability and flexibility compared to larger models.
    \item \textbf{GraphGPT.}~\cite{tang2024graphgpt} GraphGPT is a framework that integrates LLMs with graph structural knowledge through instruction tuning to enable strong generalization in graph learning tasks. It includes a text-graph grounding module and a dual-stage tuning process with a lightweight graph-text alignment projector, allowing LLMs to understand complex graph structures. GraphGPT achieves superior performance in both supervised and zero-shot graph tasks.
    \item \textbf{TAPE.}~\cite{he2023harnessing} TAPE leverages LLMs to extract textual features from TAGs by using zero-shot classification with explanations. These explanations are then interpreted into informative features for graph neural networks (GNNs), enhancing their performance on downstream tasks. The approach achieves state-of-the-art results on multiple TAG benchmarks and significantly accelerates training.
\end{itemize}
The detailed information of datasets used in this paper is introduced as follows,
\begin{itemize}
    \item \textbf{arXiv(non-text)}~\cite{hu2020open} The ogbn-arxiv dataset, part of the Open Graph Benchmark (OGB), is a citation network constructed from arXiv papers. In this dataset, nodes represent academic papers, and edges denote citation relationships between them. Each node is enriched with textual features derived from the paper’s title and abstract. The primary task associated with ogbn-arxiv is node classification, where the goal is to predict the subject area or category of each paper based on its content and citation context. This dataset is widely used for evaluating graph neural network models that integrate both structural and textual information.
    \item \textbf{Cora(non-text)}~\cite{mccallum2000automating} is a widely used citation network where nodes represent scientific papers, and edges denote citation relationships between them. Each paper is categorized into one of several predefined topics, and each node has associated features derived from the paper's content. The task commonly associated with the Cora dataset is node classification, where the objective is to predict the paper's category based on its content and citation network. It is a standard benchmark for evaluating graph-based models, particularly in the field of machine learning and network analysis.
    \item \textbf{PubMed(non-text)}~\cite{white2020pubmed} is a citation network derived from the PubMed repository of biomedical papers. Similar to Cora, nodes represent research papers, and edges indicate citation links. Each paper is categorized into one of several medical topics, with node features representing the paper’s abstract text. The primary task for the PubMed dataset is node classification, where the goal is to predict the paper's medical subject category. This dataset is widely used for evaluating graph-based models in the biomedical and healthcare domains, focusing on understanding how citations and content interact within a specific research field.
    \item \textbf{Ogbn-arxiv(text), Cora(text), Pubmed(text)} sourced from Github reopsitory provided in Chen et al~\cite{chen2024exploring}
    \item \textbf{ACM}~\cite{wang2019heterogeneous} ACM dataset is a set of extract papers published in KDD, SIGMOD, SIGCOMM, MobiCOMM, and VLDB. Wang et al divide the papers into threeclasses (Database, Wireless Communication, Data Mining). Then they construct a heterogeneous graph that comprises 3025 papers (P),5835 authors (A) and 56 subjects (S). Paper features correspond toelements of a bag-of-words represented of keywords.
    \item \textbf{DBLP}~\cite{tang2008arnetminer} DBLP is a computer science bibliography website. Fu et al adopt a subset of DBLP, containing 4057 au-thors, 14328 papers, 7723 terms, and 20 publication venuesafter data preprocessing. The authors are divided into fourresearch areas (Database, Data Mining, Artifcial Intelligence,and Information Retrieval). Each author is described by abag-of-words representation of their paper keywords.
\end{itemize}

\subsection{Results of Q2 experiment}

\begin{longtable}{cccccc}
\caption{Node Classification Performance Across Different Datasets} \label{Q2} \\

\toprule
\textbf{Model Category} & \textbf{Model} & \textbf{Dataset} & \textbf{Micro-f1} & \textbf{Macro-f1} & \textbf{AUC\_ROC} \\
\midrule
\endfirsthead

\toprule
\textbf{Model Category} & \textbf{Model} & \textbf{Dataset} & \textbf{Micro-f1} & \textbf{Macro-f1} & \textbf{AUC\_ROC} \\
\midrule
\endhead

\midrule
\multicolumn{6}{r}{\small\textit{(Continued on next page)}} \\
\endfoot

\bottomrule
\endlastfoot

\multirow{6}{*}{Homogeneous Graph Model}  & \multirow{4}{*}{GCN}       & Cora  & 0.8125 & 0.8237 & 57\%/2673\%   \\
 &      & PubMed& 0.8611 & 0.8587 & 67\%/2791\%    \\
 &      & arXiv      & 0.8621 & 0.8505 & 70\%/2764\%    \\
 &      & CITE  & 0.5163 & 0.0297 & /     \\ \cline{2-6} 
 & \multirow{2}{*}{GAT}       & Cora  & 0.8603 & 0.8488 & 59\%/1721\%     \\
 &      & PubMed& 0.8640 & 0.8493 & 60\%/1722\%    \\
 \multirow{6}{*}{Homogeneous Graph Model}
& \multirow{2}{*}{GAT}       & arXiv & 0.7179 & 0.5496 & 33\%/1079\%    \\
 &      & CITE  & 0.5395 & 0.0466 & /    \\ \cline{2-6} 
 & \multirow{4}{*}{SAGE}      & Cora  & 0.8125 & 0.8237 & 50\%/1638\%     \\
 &      & PubMed& 0.8667 & 0.8586 & 60\%/1711\%    \\
 &      & arXiv & 0.7185 & 0.5496 & 33\%/1059\%   \\
 &      & CITE  & 0.5403 & 0.0474 & /   \\ \midrule
\multirow{9}{*}{Heterogeneous Graph Model} & \multirow{3}{*}{HGT}       & DBLP  & 0.8419 & 0.8334 & -14\%/99\%    \\
 &      & ACM   & 0.8803 & 0.8762 & -10\%/109\%    \\
 &      & CITE  & 0.9594 & 0.3831 & /    \\ \cline{2-6} 
 & \multirow{3}{*}{SimpleHGN} & DBLP  & 0.9303 & 0.9245 & -6\%/83\%    \\
 &      & ACM   & 0.9063 & 0.9046 & -9\%/79\%    \\
 &      & CITE  & 0.9872 & 0.5959 & /    \\ \cline{2-6} 
 & \multirow{3}{*}{RGCN}      & DBLP  & 0.6942 & 0.5845 & -20\%/305\%    \\
 &      & ACM   & 0.7809 & 0.7567 & -9\%/425\%    \\
 &      & CITE  & 0.7484 & 0.1192 & /   \\ \midrule
\multirow{4}{*}{LLM-Centric Model}     & \multirow{4}{*}{LLAMA}     & arXiv & 0.0045 & 0.0015 & -93\%/-59\%    \\
 &      & Cora  & 0.2962 & 0.0936 & 375\%/2430\%    \\
 &      & PubMed& 0.8402 & 0.8184 & 1248\%/22019\%    \\
 &      & CITE  & 0.0623 & 0.0037 & /    \\ \midrule
\multirow{8}{*}{LLM+Graph Model}& \multirow{4}{*}{TAPE}      & arXiv & 0.7275 & 0.5123 & -27\%/-9\%    \\
 &      & Cora  & 0.7573 & 0.7401 & -24\%/31\%    \\
 &      & PubMed& 0.8022 & 0.7948 & -19\%/41\%   \\
 &      & CITE  & 0.9937 & 0.5655 & /    \\ \cline{2-6} 
 & \multirow{4}{*}{GraphGPT} & arXiv & 0.6258    & 0.2622     & 2824\%/6623\%       \\
 &      & Cora  & 0.1256    & 0.0819    & 487\%/2000\%       \\
 &      & PubMed& 0.7011    & 0.6491   & 3176\%/16544\%       \\
 &      & CITE  & 0.0214    & 0.0039    & /       \\ 

\end{longtable}

\end{document}